\documentclass{article} 
\usepackage[utf8]{inputenc}
\usepackage{iclr2026_conference,times}

\usepackage{amsmath,amsfonts,bm}

\def\eqref#1{equation~\ref{#1}}

\def\1{\bm{1}}

\DeclareMathAlphabet{\mathsfit}{\encodingdefault}{\sfdefault}{m}{sl}
\SetMathAlphabet{\mathsfit}{bold}{\encodingdefault}{\sfdefault}{bx}{n}

\usepackage{xcolor}
\usepackage{longtable}
\usepackage{tikz}
\usepackage{amsmath,bm}
\usepackage{wrapfig}
\usepackage{multirow}
\usepackage{booktabs}
\usetikzlibrary{arrows.meta,calc,shapes}
\usepackage{pgfplots}\pgfplotsset{compat=1.18}
\usepackage{hyperref}
\usepackage{url}
\usepackage{graphicx}
\usepackage{subcaption}
\usepackage{booktabs}
\usepackage{cleveref} 
\usepackage{tcolorbox} 
\tcbuselibrary{breakable} 

\usepgfplotslibrary{fillbetween} 

\usepackage{pifont} 

\usepackage{newunicodechar}

\usetikzlibrary{patterns}     

\definecolor{mainccs}{HTML}{204037}       
\definecolor{ccslight}{HTML}{4A6E63}      
\definecolor{ccssuperlight}{HTML}{84A198} 

\definecolor{palettedrab}{HTML}{B8C4C2}   
\definecolor{paletteorange}{HTML}{F5AF5A} 
\definecolor{palettegreen}{HTML}{8DA960}  
\definecolor{palettedarkrose}{HTML}{8E514F}
\definecolor{colorBalanced}{RGB}{76, 114, 176}   
\definecolor{colorSkewed}{RGB}{196, 78, 82}     
\definecolor{inkmuted}{gray}{0.45}
\definecolor{inkstrong}{gray}{0.15}
\definecolor{textDark}{RGB}{40, 40, 40}
\providecolor{clinearact}{RGB}{44,160,44}

\definecolor{wincolor}{HTML}{8DA960}  
\definecolor{tiecolor}{HTML}{F5AF5A}  
\definecolor{losecolor}{HTML}{8E514F} 

\crefname{fig.}{Fig.}{Figs.}
\Crefname{figure}{Figure}{Figures}
\crefname{tab.}{Tab.}{Tabs.}
\Crefname{table}{Table}{Tables}
\crefname{eq.}{Eq.}{Eqs.}
\Crefname{equation}{Equation}{Equations}
\crefname{sec.}{Sec.}{Secs.}
\Crefname{section}{Section}{Sections}
\AddToHook{cmd/appendix/before}{\crefalias{section}{appendix}}

\title{Contrastive Covariance Subspaces for Activation Steering}

\author{\makebox[\textwidth]{\begin{tabular}[t]{c}
\bfseries\rule{0pt}{24pt}Alexandros Xenos$^{1*\dagger}$, \quad Manos Plitsis$^{1,2*\dagger}$, \quad Niki Foteinopoulou$^{1}$, \\
\bfseries Yannis Panagakis$^{2}$, \quad Ioannis Patras$^{1,3}$ \\[4pt]
\normalfont $^{1}$Tavus.io \quad $^{2}$University of Athens \quad $^{3}$Queen Mary University of London
\end{tabular}}}

\iclrfinalcopy 
\begin{document}

\maketitle

{\renewcommand{\thefootnote}{\fnsymbol{footnote}}%
\footnotetext[1]{Equal contribution.}%
\footnotetext[2]{Corresponding authors: \texttt{\{alexandros.xenos,manos.plitsis\}@tavus.io}}}

\begin{abstract}
Activation steering controls large language model behavior by adding a direction to hidden activations during inference. For steering to work well, that direction must be a pure signal for the target concept: any confound entangled with it, such as an incidental topic or writing style, is injected into the residual stream alongside the concept itself and degrades the generated text. We introduce Contrastive Covariance Steering (CCS), a training-free method that eliminates this confound variation. CCS projects steering vectors to a low-dimensional subspace obtained in closed form by solving a contrastive variance objective, which retains the directions along which the target concept varies while suppressing directions shared with the other classes. 
Across three LLMs and four steering domains, CCS consistently beats all baseline steering methods in all 12 model–domain settings. The largest relative improvement occurs on AxBench and Emotion where we outperform the next strongest steering baseline by $48$ to $70\%$ and $13$ to $27\%$, respectively. Furthermore, CCS outperforms naive difference-of-means steering by up to $151\%$, showcasing the importance of deconfounding concepts. A toy experiment verifies our motivation on why: planting a topic nuisance into the training data reduces unprojected steering by $46\%$, whereas CCS loses only $12\%$ and recovers most of the gap to an oracle that explicitly deletes the planted direction. A human study with 23 raters on $296$ generations reproduces the LLM judge's ranking. 
Our extensive ablation verifies the importance of our method's components and robustness to the number of nuisance classes.
Finally, we extend our method to multi-turn conversations, where CCS sustains both concept expression and fluency, matching system prompt performance as other steering methods drift.

\end{abstract}
\section{Introduction}
\label{sec:introduction}

Additive activation steering offers a lightweight way to control large language model (LLM) behavior without updating model weights. By adding a concept-associated vector to intermediate activations during inference, it can influence properties such as personality, emotional tone, and truthfulness \citep{li2023inference,zou2023repe,rimsky2024caa}. Activation steering has also been explored as a safety tool~\citep{ICLR2025_e2dd5360,ICLR2025_36e3f9e6}. Although activation interventions can bypass safety-trained refusal behavior \citep{arditi2024refusal}, they can also support selective refusal and improve resistance to jailbreak attempts \citep{ICLR2025_e2dd5360,ICLR2025_36e3f9e6}.
It is effective at controlling model behavior, easy to implement and comes at no additional computational cost at inference time.

Effective control, however, requires more than increasing concept
expression. Strong interventions can move activations away from
their natural distribution and degrade generation quality
\citep{luo2026glp,soo2025interpretablesteeringlargelanguage,ICLR2025_8c3262a4}. This creates a fundamental trade-off of concept expression (\textit{how much} we enhance a given concept/behavior) and output quality. A useful steering direction should therefore strengthen the intended behavior while preserving fluent,
relevant responses.
One fundamental challenge is separating the target concept from incidental variation in its demonstrations. 
Existing methods that estimate directions using contrastive pairing \citep{turner2023activation,rimsky2024caa} can cancel shared features; systematic differences in topic, style, formatting, or response length may remain entangled with the target signal.
Approaches that incorporate second-order information can
capture dominant directions of variation,
but do not explicitly distinguish target variation from
nuisance variation \citep{zou2023repe, qiu2024spectral}. Non-additive methods
address related issues through different mechanisms like angular rotation in a 2D subspace, or optimal transport
\citep{vu_angular_2025,rodriguez2025controlling}.

\begin{figure}[t]
    \centering
    \vspace{-\baselineskip}
    \resizebox{0.6\textwidth}{!}{%
        \input{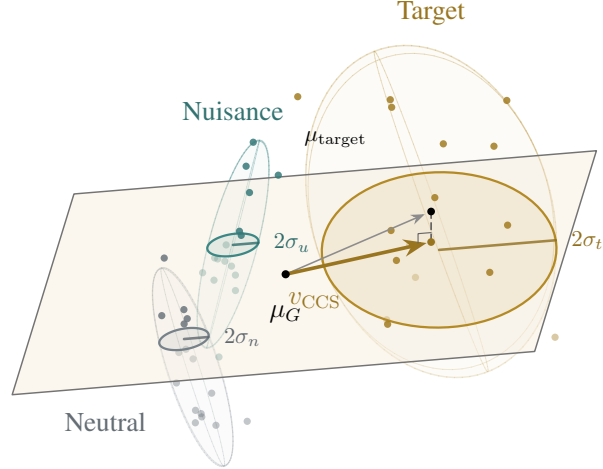}%
    }
    \caption{\textbf{CCS} projects the target--grand-mean direction
    onto a contrastive subspace, where the variance of the target class is maximised, while concurrently minimising the variances of other, neutral and nuisance, classes.}
    \label{fig:teaser-concept}
    \vspace{-0.5\baselineskip}
\end{figure}

In this paper, we introduce \textbf{Contrastive Covariance Steering (CCS)}, which \textbf{explicitly eliminates this nuisance variability} in steering directions. We optimize a contrastive subspace objective that rewards variation in the target concept while penalizing variation shared with every other available concept class, admitting a closed-form solution via eigendecomposition. CCS therefore requires no gradient-based optimization, no auxiliary trained models, and a single additive direction at one layer. Projecting a steering direction onto the resulting subspace isolates the concept's own discriminative signal from confounds such as topic, style or dataset-specific artifacts such as formatting differences or length differences. Without this projection, a naive difference-of-means direction could carry such confounds into the generated text. Figure \ref{fig:teaser-concept} shows our method's subspace construction in 3 dimensions, as well as the resulting steering vector after projection.

Our main contributions can be summarised as follows:
\begin{itemize}
\item We introduce CCS, a closed-form method that projects activations onto a contrastive subspace to isolate target concept signals from nuisance variance.

\item Our method outperformed established baselines in all model-domain settings across 3 LLMs (Llama-3.1-8B~\cite{grattafiori2024llama3herdmodels}, Qwen3-8B~\cite{yang2025qwen3technicalreport}, Gemma-2-9B~\cite{gemmateam2024gemma2improvingopen}) and 4 benchmarks. The results are verified using both LLM judge rankings and a blind multi-rater human study.

\item Ablations show that our two main mechanisms, namely grand mean centering and subspace projection, each independently improve standard difference-of-means steering.

\item In a multi-turn conversation steering experiment, CCS sustains an $80\%$ joint concept-fluency score over 20 turns and keeps near-maximum concept expression, eventually matching system prompting.

\end{itemize}

\section{Related Work}

\paragraph{Activation steering.}
Inference-time activation steering controls LLM behavior by modifying internal representations without updating model weights
\citep{subramani2022extracting,turner2023activation,zou2023repe,li2023inference}.
A common approach derives a single direction from contrasting activations, as in Contrastive Activation Addition (CAA; \citealp{rimsky2024caa}), with related work exploring mean-centering \citep{jorgensen2023improving}, persona-specific directions \citep{chen_persona_2025}, affine transformations \citep{singh2024representation,rodriguez2025controlling}, rotations \citep{vu_angular_2025}, and learned interventions \citep{wu2024reft,wu2025improved,bao2026faithful}.
CCS remains in the closed-form additive setting, but constructs steering directions from mean-centered activations, using second-order information from multiple competing concepts.

\paragraph{Spectral approaches and low-rank subspaces.}
CCS is closely related to spectral methods that identify low-dimensional structure in representation spaces.
PCA and contrastive PCA identify directions of high variance, with cPCA emphasizing variation present in a target dataset but not a background dataset
\citep{Pearson01111901,abid2018cpca}; \citet{NEURIPS2023_07cf32cf} extend this principle to multiple competing factors.
In LLMs, SEA uses spectral transformations derived from positive and negative activation structure \citep{qiu2024spectral}, conceptor-based steering models concepts as multidimensional subspaces using second-order statistics \citep{triantafyllopoulos2026conceptorssemanticsteering}, and \citet{bhandari_activation-space_2026} project personality directions through a shared low-rank subspace.
CCS instead learns a \emph{target-specific contrastive} subspace that rewards target scatter and penalizes non-target scatter, then uses it to filter the target--grand-mean contrast into a single additive steering vector.
This is also related to Fisher Discriminant Analysis \citep{fisher1936lda}, although FDA learns a common discriminative subspace, whereas CCS learns a separate subspace for each target.
Recent work further shows that behavioral representations occupy overlapping low-rank subspaces and orthogonalization alone does not guarantee behavioral independence
\citep{bhandari2026personalitytraitsinterferegeometric,sharma2026lowranksubspaceanalysisllm}.

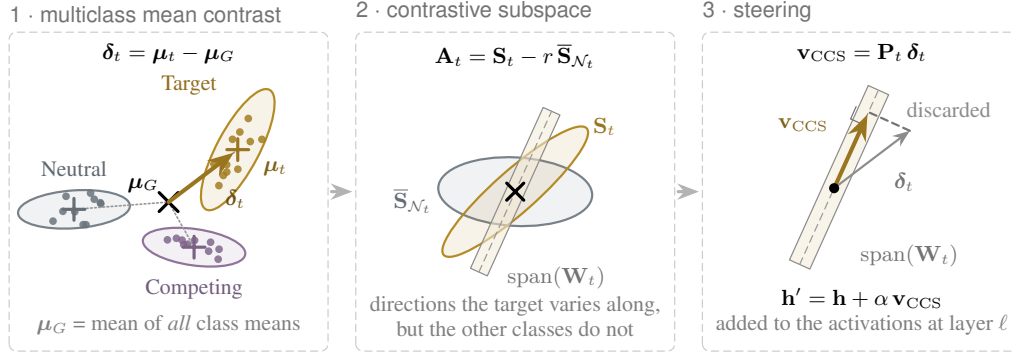
\begin{figure}[t]
\centering
\resizebox{\textwidth}{!}{
\begingroup
\definecolor{target}{HTML}{B48A24}
\definecolor{neutral}{HTML}{77818A}
\definecolor{competing}{HTML}{8F739B}

\tikzset{
  pnl/.style   ={draw=black!20,densely dashed,rounded corners=3pt,line width=.5pt},
  hdr/.style   ={font=\small\sffamily,text=black!55,inner sep=0pt},
  eqn/.style   ={font=\small,inner sep=0pt},
  tny/.style   ={font=\footnotesize,inner sep=1.5pt},
  cloud/.style ={fill=#1,fill opacity=.10,draw=#1,line width=.9pt},
  dotc/.style  ={fill=#1!90!black,opacity=.85},
  sv/.style    ={->,>=Stealth,line width=1.7pt,#1},
  tv/.style    ={->,>=Stealth,black!45,line width=.9pt},
  pln/.style   ={fill=yellow!22!brown!20,fill opacity=.55,draw=black!45,line width=.5pt},
}
\newcommand{\cmean}[3]{\draw[#1,line width=1.3pt,line cap=round]
  (#2,#3)++(-.15,0)--++(.30,0) (#2,#3)++(0,-.15)--++(0,.30);}
\newcommand{\gmean}[3]{\draw[#1,line width=1.3pt,line cap=round]
  (#2,#3)++(-.13,-.13)--++(.26,.26) (#2,#3)++(-.13,.13)--++(.26,-.26);}

\begin{tikzpicture}[line cap=round,line join=round,>=Stealth]
\def\pw{4.60}\def\ph{4.60}\def\gp{0.40}

\begin{scope}[shift={(0,0)}]
  \draw[pnl] (0,0) rectangle (\pw,\ph);
  \node[hdr,anchor=south west] at (0,\ph+.14) {1 $\cdot$ multiclass mean contrast};
  \node[eqn,anchor=north] at (\pw/2,\ph-.16)
    {$\boldsymbol\delta_t=\boldsymbol\mu_t-\boldsymbol\mu_G$};

  \begin{scope}[shift={(2.30,2.15)},scale=1.05]
    \begin{scope}[shift={(0.95,0.72)},rotate=62]
      \draw[cloud=target] (0,0) ellipse (0.93 and 0.29);\end{scope}
    \begin{scope}[shift={(-1.30,-0.10)},rotate=4]
      \draw[cloud=neutral] (0,0) ellipse (0.77 and 0.26);\end{scope}
    \begin{scope}[shift={(0.35,-0.62)},rotate=-8]
      \draw[cloud=competing] (0,0) ellipse (0.71 and 0.25);\end{scope}
    \foreach \p in {(0.691,0.226),(1.122,0.831),(0.799,0.413),(0.979,0.962),
                    (0.752,0.365),(0.878,0.862),(0.951,0.451),(1.282,0.861),
                    (1.095,1.148),(0.798,0.513),(0.622,0.524)}
      \fill[dotc=target] \p circle (1.5pt);
    \foreach \p in {(-1.134,-0.311),(-1.022,0.039),(-0.931,-0.024),(-1.289,-0.134),
                    (-0.938,-0.009),(-1.452,-0.127),(-1.174,-0.310),(-1.072,0.122),
                    (-1.588,-0.262),(-1.409,0.079)}
      \fill[dotc=neutral] \p circle (1.5pt);
    \foreach \p in {(0.185,-0.523),(0.107,-0.498),(0.262,-0.579),(0.693,-0.648),
                    (0.577,-0.552),(-0.016,-0.519),(0.208,-0.585),(0.550,-0.675),
                    (0.357,-0.527),(0.681,-0.785)}
      \fill[dotc=competing] \p circle (1.5pt);

    \draw[densely dotted,black!40,line width=.7pt] (0.95,0.72)--(0,0);
    \draw[densely dotted,black!40,line width=.7pt] (-1.30,-0.10)--(0,0);
    \draw[densely dotted,black!40,line width=.7pt] (0.35,-0.62)--(0,0);

    \cmean{target!85!black}{0.95}{0.72}
    \cmean{neutral!85!black}{-1.30}{-0.10}
    \cmean{competing!85!black}{0.35}{-0.62}
    \gmean{black}{0}{0}
    \node[tny,anchor=south east] at (-0.08,0.08) {$\boldsymbol\mu_G$};
    \node[tny,text=target!70!black,anchor=west] at (1.24,0.46) {$\boldsymbol\mu_t$};

    \draw[sv=target!85!black,shorten >=2pt] (0,0)--(0.95,0.72);
    \node[tny,text=target!60!black,anchor=north west] at (0.72,0.22) {$\boldsymbol\delta_t$};

    \node[tny,text=target!80!black,anchor=south]   at (0.28,1.42) {Target};
    \node[tny,text=neutral!80!black,anchor=south]  at (-1.30,0.30) {Neutral};
    \node[tny,text=competing!80!black,anchor=north]at (0.35,-1.00) {Competing};
  \end{scope}
  \node[tny,text=black!50,anchor=south] at (\pw/2,0.18)
    {$\boldsymbol\mu_G$ = mean of \emph{all} class means};
\end{scope}

\begin{scope}[shift={(\pw+\gp,0)}]
  \draw[pnl] (0,0) rectangle (\pw,\ph);
  \node[hdr,anchor=south west] at (0,\ph+.14) {2 $\cdot$ contrastive subspace};
  \node[eqn,anchor=north] at (\pw/2,\ph-.16)
    {$\mathbf A_t=\mathbf S_t-r\,\overline{\mathbf S}_{\mathcal N_t}$};

  \begin{scope}[shift={(2.30,2.30)},scale=.95]
    \begin{scope}[rotate=-3.1]
      \draw[cloud=neutral] (0,0) ellipse (1.17 and 0.52);\end{scope}
    \begin{scope}[rotate=41.5]
      \draw[cloud=target] (0,0) ellipse (1.45 and 0.32);\end{scope}
    \begin{scope}[rotate=65.69]
      \draw[pln] (-1.34,-0.14) rectangle (1.34,0.14);
      \draw[densely dashed,black!40,line width=.5pt] (-1.34,0)--(1.34,0);\end{scope}
    \gmean{black}{0}{0}
    \node[tny,text=target!85!black,anchor=west]  at (1.12,0.99) {$\mathbf S_t$};
    \node[tny,text=neutral!85!black,anchor=east] at (-1.20,-0.14) {$\overline{\mathbf S}_{\mathcal N_t}$};
    \node[tny,text=black!55,anchor=west] at (-0.18,-1.30) {$\mathrm{span}(\mathbf W_t)$};
  \end{scope}
  \node[tny,text=black!50,anchor=south,align=center] at (\pw/2,0.16)
    {directions the target varies along,\\but the other classes do not};
\end{scope}

\begin{scope}[shift={(2*\pw+2*\gp,0)}]
  \draw[pnl] (0,0) rectangle (\pw,\ph);
  \node[hdr,anchor=south west] at (0,\ph+.14) {3 $\cdot$ steering};
  \node[eqn,anchor=north] at (\pw/2,\ph-.16)
    {$\mathbf v_{\mathrm{CCS}}=\mathbf P_t\,\boldsymbol\delta_t$};

  \begin{scope}[shift={(1.90,2.35)},scale=1.15]
    \begin{scope}[rotate=65.69]
      \draw[pln] (-1.05,-0.14) rectangle (1.35,0.14);
      \draw[densely dashed,black!40,line width=.5pt] (-1.05,0)--(1.35,0);\end{scope}
    \node[tny,text=black!55,anchor=west] at (0.28,-0.85) {$\mathrm{span}(\mathbf W_t)$};

    \draw[tv] (0,0)--(0.95,0.72);
    \node[tny,text=black!55,anchor=north west] at (0.70,0.24) {$\boldsymbol\delta_t$};
    \draw[densely dashed,black!55,line width=.9pt] (0.95,0.72)--(0.431,0.954);
    \node[tny,text=black!45,anchor=west] at (0.86,0.98) {discarded};
    \begin{scope}[shift={(0.431,0.954)},rotate=65.69]
      \draw[black!55,line width=.5pt] (-0.17,0)--(-0.17,0.17)--(0,0.17);\end{scope}
    \draw[sv=target!85!black,shorten >=1pt] (0,0)--(0.431,0.954);
    \node[tny,text=target!70!black,anchor=east] at (-0.02,0.80) {$\mathbf v_{\mathrm{CCS}}$};
    \fill[black] (0,0) circle (1.8pt);
  \end{scope}

  \node[eqn,anchor=south] at (\pw/2,0.62)
    {$\mathbf h'=\mathbf h+\alpha\,\mathbf v_{\mathrm{CCS}}$};
  \node[tny,text=black!50,anchor=south] at (\pw/2,0.18)
    {added to the activations at layer $\ell$};
\end{scope}

\foreach \i in {0,1}{
  \pgfmathsetmacro\ax{(\i+1)*\pw+\i*\gp+0.05}
  \draw[->,>=Stealth,black!30,line width=1.3pt] (\ax,\ph/2)--(\ax+\gp-0.10,\ph/2);}
\end{tikzpicture}
\endgroup}
\caption{\textbf{Contrastive Covariance Steering.}
\textbf{(1)}~The steering direction is measured from the grand mean of \emph{all}
class means, not from a single neutral reference.
\textbf{(2)}~$\mathbf{A}_t=\mathbf{S}_t-r\,\overline{\mathbf{S}}_{\mathcal{N}_t}$ rewards target scatter and penalises the mean
non-target scatter $\overline{\mathbf{S}}_{\mathcal{N}_t}=|\mathcal{N}_t|^{-1}\sum_{c\in\mathcal{N}_t}\mathbf{S}_c$ (Eq.~\ref{eq:contrastive_scatter}); its top-$k$ eigenvectors span directions along which the target
varies but the other classes do not. 
\textbf{(3)}~Projecting $\boldsymbol{\delta}_t$ onto that subspace discards the
shared component.}
\label{fig:method}
\end{figure}

\section{Method}
\label{sec:method}

In this section, we introduce Contrastive Covariance Steering (CCS), a closed-form method combining multiclass mean contrasts with contrastive subspace projection to construct additive steering directions. Real activation spaces are more entangled than a simple contrast between two class means can capture: pushing toward a target concept often drags the model along nuisance directions shared with other classes in the dataset. A simple difference of means can encode irrelevant information such as variability in topic, writing style or length, especially if the dataset has not controlled for these confounds by explicitly balancing on confounds (even then, implicit imbalances can still remain). CCS addresses this by first measuring the target's displacement from the mean of all class means rather than a single neutral reference class, stabilizing the first-order displacement.  It then projects that displacement onto a subspace that preserves the target's own variation while suppressing variation shared with the other classes. Our method is outlined end-to-end in~\Cref{fig:method}.

\subsection{Background}

We first formalize the mathematical notation for residual-stream activations and standard additive steering interventions. 

Let $\mathbf{h}_{\ell,s}(\mathbf{x})\in\mathbb{R}^d$ denote the
residual-stream activation of a language model with prompt $x$, at layer $\ell$ and
token position $s$. At a given layer, we use the
mean-pooled prompt activations,
$\mathbf{h}(\mathbf{x})=T^{-1}\sum_{s=1}^{T}
\mathbf{h}_{\ell,s}(\mathbf{x})$.

An \textit{additive activation steering} method finds a vector $\mathbf{v}$ in the model's activation space, then adds these vectors to the residual stream activations of each token during generation, after unit-normalizing and multiplying by a \textit{steering strength} factor to control intensity: 
\begin{equation}
    \mathbf{h}_{\ell,s}^{\,\mathrm{steer}}
    =
    \mathbf{h}_{\ell,s}
    +
    \alpha\widehat{\mathbf{v}}.
    \label{eq:normalized_steering}
\end{equation}
where $\widehat{\mathbf{v}} = \mathbf{v} / \|\mathbf{v}\|_2$. We don't use a constant $\alpha$ when steering, but a factor of each steered token's norm (see Appendix~\ref{app:per-token-norm}).

Given concept classes $c\in\mathcal{C}$, we collect the activations of the
prompts belonging to each class, $\mathcal{H}_c=\{\mathbf{h}_{c,i}\}_{i=1}^{n_c}$,
with class mean $\boldsymbol{\mu}_c=n_c^{-1}\sum_i\mathbf{h}_{c,i}$, and
center every activation on the mean of the $|\mathcal{C}|$ class
means:
\begin{equation}
    \boldsymbol{\mu}_G
    =\frac{1}{|\mathcal{C}|}\sum_{c\in\mathcal{C}}\boldsymbol{\mu}_c,
    \qquad
    \widetilde{\mathbf{h}}_{c,i}
    =\mathbf{h}_{c,i}-\boldsymbol{\mu}_G,
    \qquad
    \boldsymbol{\delta}_c=\boldsymbol{\mu}_c-\boldsymbol{\mu}_G .
    \label{eq:grand_mean}
\end{equation}

The scatter of class $c$ about this common center is
\begin{equation}
    \mathbf{S}_c
    =\widetilde{\mathbf{H}}_c\widetilde{\mathbf{H}}_c^{\top}
    =\frac{1}{n_c}\sum_{i=1}^{n_c}
      \widetilde{\mathbf{h}}_{c,i}\widetilde{\mathbf{h}}_{c,i}^{\top}
    =\boldsymbol{\Sigma}_c+\boldsymbol{\delta}_c\boldsymbol{\delta}_c^{\top},
    \label{eq:class_scatter}
\end{equation}
where $\boldsymbol{\Sigma}_c$ is the within-class covariance. Unlike
$\boldsymbol{\Sigma}_c$ alone, $\mathbf{S}_c$ retains both the class's internal
variation and its displacement from the grand mean, so a direction is penalised
either because a nuisance class varies along it or because that class sits away
from the origin along it.

\paragraph{Contrastive subspace.}
Let $\widetilde{\mathbf{H}}_c\in\mathbb{R}^{d\times n_c}$ contain in its columns
the centered activations $\mathbf{h}-\boldsymbol{\mu}_G$ of class $c$, scaled
by $n_c^{-1/2}$ so that $\|\mathbf{W}^\top\widetilde{\mathbf{H}}_c\|_F^2$ is
the mean squared norm of the class's activations after projection onto
$\mathbf{W}$, independent of class size. For a target class $t$ with nuisance
classes $\mathcal{N}_t=\mathcal{C}\setminus\{t\}$, we seek a $k$-dimensional
subspace in which the target's activations have large norm while those of the
nuisance classes are close to zero:
\begin{equation}
    \mathbf{W}_t=
    \underset{\mathbf{W}^\top\mathbf{W}=\mathbf{I}_k}{\operatorname{arg\,max}}
    \bigg\{
      \big\|\mathbf{W}^\top\widetilde{\mathbf{H}}_t\big\|_F^2
      -\frac{r}{|\mathcal{N}_t|}\sum_{c\in\mathcal{N}_t}
      \big\|\mathbf{W}^\top\widetilde{\mathbf{H}}_c\big\|_F^2
    \bigg\},
    \label{eq:ccs-objective}
\end{equation}
where the contrastive weight $r\ge0$ sets the importance of suppressing
variation in a typical nuisance class relative to preserving variation in the
target class. Since
$\|\mathbf{W}^\top\widetilde{\mathbf{H}}_c\|_F^2
=\operatorname{tr}(\mathbf{W}^\top\widetilde{\mathbf{H}}_c\widetilde{\mathbf{H}}_c^\top\mathbf{W})
=\operatorname{tr}(\mathbf{W}^\top\mathbf{S}_c\mathbf{W})$,
Eq.~\ref{eq:ccs-objective} is the trace maximization
\begin{equation}
    \mathbf{W}_t=
    \underset{\mathbf{W}^\top\mathbf{W}=\mathbf{I}_k}{\operatorname{arg\,max}}\;
    \operatorname{tr}\!\big(\mathbf{W}^\top\mathbf{A}_t\mathbf{W}\big),
    \qquad
    \mathbf{A}_t=\mathbf{S}_t-\frac{r}{|\mathcal{N}_t|}\sum_{c\in\mathcal{N}_t}\mathbf{S}_c ,
    \label{eq:contrastive_scatter}
\end{equation}
whose solution is given in closed form by the $k$ leading eigenvectors of the
symmetric matrix $\mathbf{A}_t$ \citep[Cor.~4.3.39]{horn2012matrix}. Subspace
estimation therefore requires a single eigendecomposition and no gradient-based
optimization.

\paragraph{Steering-vector construction.}
Writing the leading eigenvectors as the columns of $\mathbf{W}_t$ and
$\mathbf{P}_t=\mathbf{W}_t\mathbf{W}_t^{\top}$ for the corresponding orthogonal
projector, we project the target's displacement onto the contrastive subspace:
\begin{equation}
    \mathbf{v}_t
    =\mathbf{P}_t\boldsymbol{\delta}_t
    =\mathbf{P}_t(\boldsymbol{\mu}_t-\boldsymbol{\mu}_G).
    \label{eq:steering_vector}
\end{equation}
This results in our additive steering direction. Geometrically, $\mathbf{P}_t$ acts as a filter tailored to the target: it
removes from $\boldsymbol{\delta}_t$ the components lying along directions in
which the nuisance classes, on average, scatter more than the target does, and
retains the components specific to the target. 

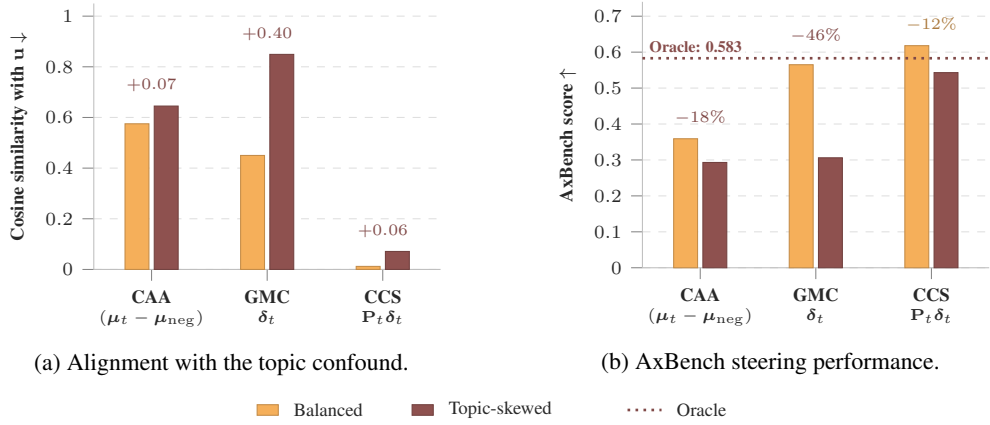
\begin{figure*}[t]
\centering
\begingroup
\pgfplotsset{planted diagnostic/.style={
    ybar, bar width=9pt,
    width=0.92\linewidth, height=5.1cm,
    scale only axis=false,
    xmin=0.5, xmax=3.5, enlarge x limits=false,
    xtick={1,2,3},
    xticklabel style={font=\scriptsize,align=center,text=textDark},
    xticklabels={
        {\textbf{CAA}\\[-1pt]{\tiny $(\boldsymbol{\mu}_t-\boldsymbol{\mu}_{\mathrm{neg}})$}},
        {\textbf{GMC}\\[-1pt]{\tiny $\boldsymbol{\delta}_t$}},
        {\textbf{CCS}\\[-1pt]{\tiny $\mathbf P_t\boldsymbol{\delta}_t$}}
    },
    ylabel style={font=\scriptsize\bfseries,text=textDark},
    yticklabel style={font=\scriptsize,text=textDark},
    axis line style={gray!50},
    axis x line*=bottom,axis y line*=left,
    ymajorgrids=true,grid style={dashed,gray!25},
    tick align=outside,clip=false
}}
\begin{subfigure}[t]{0.48\textwidth}
\centering
\vspace{0pt}
\begin{tikzpicture}
\begin{axis}[planted diagnostic,
    ylabel={Cosine similarity with $\mathbf u$ $\downarrow$},
    ymin=0,ymax=1.05,ytick={0,0.2,0.4,0.6,0.8,1.0}]
\addplot[fill=paletteorange,draw=paletteorange!80!black]
    coordinates {(1,0.575) (2,0.450) (3,0.012)};
\addplot[fill=palettedarkrose,draw=palettedarkrose!80!black]
    coordinates {(1,0.645) (2,0.849) (3,0.071)};
\node[font=\tiny\bfseries,text=palettedarkrose!90!black,anchor=south]
    at (axis cs:1,0.665) {$+0.07$};
\node[font=\tiny\bfseries,text=palettedarkrose!90!black,anchor=south]
    at (axis cs:2,0.869) {$+0.40$};
\node[font=\tiny\bfseries,text=palettedarkrose!90!black,anchor=south]
    at (axis cs:3,0.091) {$+0.06$};
\end{axis}
\end{tikzpicture}
\caption{Alignment with the topic confound.}
\label{fig:planted-cos-sim}
\end{subfigure}%
\hfill
\begin{subfigure}[t]{0.48\textwidth}
\centering
\vspace{0pt}
\begin{tikzpicture}
\begin{axis}[planted diagnostic,
    ylabel={AxBench score $\uparrow$},
    ymin=0,ymax=0.74,ytick={0,0.1,0.2,0.3,0.4,0.5,0.6,0.7}]
\addplot[fill=paletteorange,draw=paletteorange!80!black]
    coordinates {(1,0.359) (2,0.565) (3,0.618)};
\addplot[fill=palettedarkrose,draw=palettedarkrose!80!black]
    coordinates {(1,0.293) (2,0.306) (3,0.543)};
\draw[palettedarkrose!90!black,line width=1pt,dotted]
    (axis cs:0.5,0.583)--(axis cs:3.5,0.583);
\node[font=\tiny\bfseries,text=palettedarkrose!95!black,
    fill=white,fill opacity=.9,text opacity=1,inner sep=1.2pt,
    anchor=south west] at (axis cs:0.53,0.588) {Oracle: 0.583};
\node[font=\tiny\bfseries,text=palettedarkrose!90!black,anchor=south]
    at (axis cs:1,0.373) {$-18\%$};
\node[font=\tiny\bfseries,text=palettedarkrose!90!black,anchor=south]
    at (axis cs:2,0.602) {$-46\%$};
\node[font=\tiny\bfseries,text=paletteorange!70!black,anchor=south]
    at (axis cs:3,0.632) {$-12\%$};
\end{axis}
\end{tikzpicture}
\caption{AxBench steering performance.}
\label{fig:planted-axb-score}
\end{subfigure}

\par\vspace{1ex}
\begin{tikzpicture}[font=\scriptsize,text=textDark]
\filldraw[fill=paletteorange,draw=paletteorange!80!black]
    (0,0) rectangle (.28,.14);
\node[anchor=west] at (.36,.07) {Balanced};
\filldraw[fill=palettedarkrose,draw=palettedarkrose!80!black]
    (2.05,0) rectangle (2.33,.14);
\node[anchor=west] at (2.41,.07) {Topic-skewed};
\draw[palettedarkrose!90!black,line width=1pt,dotted]
    (4.9,.07)--(5.35,.07);
\node[anchor=west] at (5.43,.07) {Oracle};
\end{tikzpicture}
\endgroup
\caption{\textbf{Planted nuisance diagnostic on AxBench.}
(a) Cosine similarity between the steering vector and the planted topic direction
$u$. Under topic skew, GMC reaches 0.849, compared with 0.071 for CCS.
(b) Topic skew reduces GMC's AxBench score by 46\%, whereas CCS retains
88\% of its balanced score without topic labels. Oracle performance is included for comparison, obtained by explicitly removing $u$ via $\delta_t-u(u^\top\delta_t)$}.
\label{fig:planted-nuisance-combined}
\end{figure*}

\subsection{Does Projection Suppress Topic Confounds?}
\label{sec:toy-projection}

To isolate how latent dataset confounds corrupt steering vectors and test whether contrastive projection mitigates this vulnerability, we design a toy experiment on ten AxBench concepts \citep{wu2025axbench}. We hypothesize that (1) unprojected mean shifts are highly sensitive to topic skew, and (2) contrastive subspace projection restores robustness by filtering out shared nuisance variance. To test this, we cluster instruction activation centroids into five topic groups and construct equal-sized datasets: a \emph{balanced} set preserving natural topic proportions and a \emph{topic-skewed} set with approximately 90\% of examples from one cluster.

We compare difference-of-means (CAA), grand-mean centering (GMC), \emph{Oracle}, and \textbf{CCS}. Oracle removes the component along the unit topic direction $u_g$ using $\delta_t-u_g(u_g^\top\delta_t)$ and ground-truth topic annotations. CCS instead projects using concept-labeled scatter, without topic labels.

Under topic skew, GMC's AxBench score falls by 46\% (0.565 to 0.306), while Oracle reaches 0.583 (\Cref{fig:planted-nuisance-combined}). CCS reaches 0.543, retaining 88\% of its balanced-condition performance. The vector geometry supports the proposed mechanism: GMC's cosine similarity with $u_g$ reaches 0.849, whereas CCS remains nearly orthogonal at 0.071. Together, these results indicate contrastive projection improves robustness by suppressing topic-aligned contamination without explicit topic supervision.

\section{Experiments}
\label{sec:experiments}

\subsection{Experimental Setup}
\label{sec:experimental-setup}

\paragraph{Benchmarks and metrics.} We evaluate across four distinct domains: \textbf{AxBench} ~\cite{wu2025axbench}, a benchmark for LLM steering containing multiple open-ended concepts across different topics, and a shared \textit{negative} (concept-free) set. \textbf{BIG5 / NPTI}~\cite{li-etal-2025-big5,kang_prompt-activation_2026} includes the Big Five personality traits (openness, conscientiousness, extraversion, agreeableness and neuroticism); steering directions are extracted from BIG5-Chat~\cite{li-etal-2025-big5} and evaluated on NPTI's held-out prompts. \textbf{Emotion}~\cite{sofroniew2026emotion} includes the Ekman's \citep{Ekman1993-EKMFEA} basic emotions. \textbf{TruthfulQA} ~\cite{lin-etal-2022-truthfulqa} contains examples of truthful and untruthful sentences. 

All domains are evaluated using LLM as a judge.\footnote{We use Qwen-2.5-32B-Instruct for all domains except TQA.} For TruthfulQA we use the original paper's metrics of Truthfulness and Informativeness, evaluated by two dedicated fine-tuned Llama-2-7B judge models. We report their product, $T^*I=\text{True\%} \times \text{Info\%}/100$. For AxBench, we report the harmonic mean of \emph{concept steering effectiveness}, \emph{instruction-following}, and \emph{fluency}, rated (0--2)~\citet{wu2025axbench}.
For Emotion and BIG5 we evaluate on the same axes, using a $0-100$ scale. We report the max effectiveness when fluency and prompt adherence are both above an $85\%$ threshold, ensuring that steered text is sufficiently fluent. In Appendix \ref{app:gate-sensitivity} we show results for different thresholds, as well as pareto fronts for AxBench, BIG5, and Emotion experiments.

\paragraph{Baselines.} We compare against six activation-steering methods:
\textbf{CAA}~\cite{rimsky2024caa}: Standard difference-of-means direction.
\textbf{RepE}~\cite{zou2023repe}: Top right singular vector of $X_{\text{high}} - \mu_{\text{low}}$, sign-aligned with the mean difference. \textbf{PCA}~\cite{Pearson01111901}: top-k principal components of the target class's own (own-mean-centred) activations.
\textbf{SEA}~\cite{qiu2024spectral}: Forms cross-covariances from positive and negative activation structure, keeping directions aligned with the target class while discarding those aligned with the reference class.
\textbf{Linear-AcT}~\cite{rodriguez2025controlling}: Solves the 1-D optimal-transport problem between the reference and target marginals and fits an affine map, applied elementwise at inference.
\textbf{Angular Steering}~\cite{vu_angular_2025}: Rotates activations within a fixed
two-dimensional plane. \emph{Prompting} induces the same target concept purely via a short instruction in the prompt (e.g. "Focus on incorporating \{concept\} throughout your responses."). 
We also report the generations of the unsteered models.


\paragraph{LLMs.} We construct steering vectors for three instruction-tuned models: Llama-3.1-8B-Instruct~\cite{grattafiori2024llama3herdmodels}, Qwen3-8B~\cite{yang2025qwen3technicalreport}, and Gemma-2-9B-Instruct~\cite{gemmateam2024gemma2improvingopen}.
We use a fixed intervention layer for each model: layer 15 for Llama-3.1-8B-Instruct and 21 for Qwen3-8B and Gemma-2-9B-Instruct. This id corroborated by prior findings of effective steering layers shared across multiple
behaviors \citep{tan2024steering,rimsky2024caa}.

\begin{table*}[t]
\centering
\caption{\textbf{Steering performance across three LLMs and four benchmarks.} Reported metrics show average ratings per concept. Scores range [0, 2] on AxB and from 0--100 on others; higher is better. CCS outperforms all steering baselines in all three models.}
\label{tab:steering_results}
\vspace{2pt}
\resizebox{\textwidth}{!}{%
\small
\setlength{\tabcolsep}{2.5pt}
\begin{tabular}{lcccccccccccc}
\toprule
\multirow{2}{*}{Method} & \multicolumn{4}{c}{Llama-3.1-8B-Instruct} & \multicolumn{4}{c}{Qwen3-8B} & \multicolumn{4}{c}{Gemma-2-9B-Instruct} \\
\cmidrule(lr){2-5} \cmidrule(lr){6-9} \cmidrule(lr){10-13}
 & T*I & AxB & BIG5 & Emo. & T*I & AxB & BIG5 & Emo. & T*I & AxB & BIG5 & Emo. \\
\midrule
Unsteered & 51.8 & 0.16 & 66.9 & 16.5 & 56.2 & 0.15 & 67.7 & 19.7 & 79.5 & 0.17 & 60.7 & 23.2 \\
Prompting & 58.7 & 0.97 & 75.2 & 74.3 & 52.6 & 1.05 & 85.8 & 66.3 & 71.9 & 1.03 & 70.4 & 76.7 \\
\midrule
CAA & 72.8 & 0.31 & 75.2 & 58.6 & 63.7 & 0.41 & 77.2 & 61.9 & 84.8 & 0.48 & 77.2 & 78.9 \\
RepE & 57.5 & 0.31 & 68.8 & 56.4 & 51.4 & 0.33 & 71.8 & 60.7 & 79.9 & 0.44 & 75.8 & 78.4 \\
PCA & 55.3 & 0.46 & 79.6 & 64.7 & 52.0 & 0.49 & 81.2 & 63.1 & 85.7 & 0.53 & 72.7 & 77.5 \\
SEA & 53.3 & 0.24 & 67.8 & 17.7 & 51.9 & 0.23 & 68.3 & 20.9 & 79.7 & 0.24 & 61.7 & 24.8 \\
Linear-AcT & 58.5 & 0.40 & 82.9 & 50.0 & 57.2 & 0.58 & 81.6 & 50.9 & 81.6 & 0.48 & 76.7 & 51.6 \\
Angular Steering & 56.5 & 0.27 & 68.2 & 18.1 & 54.6 & 0.29 & 70.2 & 22.3 & 82.1 & 0.25 & 62.8 & 27.2 \\
\midrule
CCS $(k=32,r=1)$ & 79.9 & 0.61 & 80.1 & 70.4 & 75.4 & 0.67 & 84.2 & 77.2 & 89.3 & 0.70 & 76.5 & 88.7 \\
CCS ($k=32$) & 79.9 & 0.75 & 81.2 & 76.6 & 75.4 & 0.83 & 85.9 & 78.3 & 90.2 & 0.82 & 80.7 & 89.2 \\
CCS & \textbf{80.3} & \textbf{0.78} & \textbf{83.1} & \textbf{77.9} & \textbf{76.8} & \textbf{0.86} & \textbf{86.3} & \textbf{80.1} & \textbf{92.2} & \textbf{0.84} & \textbf{82.4} & \textbf{89.4} \\
\bottomrule
\end{tabular}%
}
\end{table*}

\paragraph{Hyperparameter selection.}
For each method, we tune hyperparameters separately for each concept, using 3-fold cross-validation across evaluation prompts, with the final parameters selected by majority voting. 
For CCS, we additionally report the method's performance for a fixed $k$ and both $k$ and $r$ fixed, as a good default configuration.
We evaluate all methods on the same strength grid, choosing the strength with the maximum score. We report each domain's score as the unweighted mean of its concept scores. Details about the hyperparameter selection can be found in~\Cref{app:hparams-selection}.

\subsection{Steering Vector Comparison}
\label{sec:main_experiment}

\Cref{tab:steering_results} shows results for our main experiment. CCS's gains extend
across topical, factual, and behavioral domains. 
Fully tuned CCS outperforms every activation baseline
in all 12 settings, with the clearest win on AxBench,
where gains over the strongest baseline range from $0.28$ to $0.32$ ($48-70\%$) points, while gains over standard difference of means steering from $0.36$ to $0.47$ points ($75-151\%$).
Notably, CCS achieves these results with a single additive direction at one layer, whereas Linear-AcT, which follows closely in several settings, transforms activations across multiple layers. In Appendix \ref{app:gate-sensitivity} and \ref{app:axb-gate}, results show that our method's lead over the baselines is robust across fluency and adherence cutoff points, effectively minimizing the trade-off between steering effectiveness and output quality.
Prompting remains strongest on AxBench, suggesting that explicit instructions are particularly effective for topical control. CCS instead leads prompting on TruthfulQA, BIG5, and Emotion across all three models. 

The advantages of our method are visible even with \textit{minimal hyperparameter tuning}. Uniformly setting
$k=32$ still beats every non-CCS activation baseline in
11 of 12 settings; additionally, fixing both $(k,r)=(32,1)$ does so in 10 tasks. The shared configuration is especially effective on TruthfulQA, even matching the $k=32$ CCS variant on Llama and Qwen. Concept-specific tuning matters more on AxBench, where it improves scores by $0.14-0.19$ points over the shared configuration. Thus, a common default provides a strong starting point, while tuning of hyperparameters delivers task-dependent gains.

\begin{figure*}[t]
\centering
\begingroup
\pgfplotsset{human study axis/.style={
    width=\linewidth,height=5.8cm,
    scale only axis=false,
    axis x line*=bottom,axis y line*=left,
    axis line style={black!30},
    tick style={draw=none},
    ymajorgrids=true,grid style={black!10},
    ylabel style={font=\small,text=textDark},
    yticklabel style={font=\scriptsize,text=textDark},
    xticklabel style={font=\scriptsize,text=textDark,align=center},
    xticklabel shift=3pt,
    legend style={
        at={(0.5,1.04)},anchor=south,
        draw=none,fill=none,font=\scriptsize,
        /tikz/every even column/.append style={column sep=5pt}},
    legend image code/.code={
        \draw[##1] (0cm,-.07cm) rectangle (.22cm,.07cm);},
    clip=false
}}
\begin{subfigure}[t]{0.52\textwidth}
\centering
\vspace{0pt}
\begin{tikzpicture}
\begin{axis}[ybar=6pt,bar width=10pt,human study axis,
    ylabel={AxBench score},
    ymin=0,ymax=1.2,ytick={0,.5,1},
    xmin=.5,xmax=5.5,enlarge x limits=false,
    xtick={1,2,3,4,5},
    xticklabels={CCS,{Linear-\\AcT},PCA,CAA,Unsteered},
    legend columns=2,
    nodes near coords={\pgfmathprintnumber[fixed,fixed zerofill,precision=2]
        {\pgfplotspointmeta}},
    every node near coord/.append style={
        font=\tiny,text=textDark,anchor=south,inner sep=1.2pt}
]
\addplot[fill=paletteorange,draw=paletteorange!75!black]
    coordinates {(1,1.01) (2,.50) (3,.47) (4,.40) (5,.20)};
\addplot[fill=palettegreen,draw=palettegreen!75!black]
    coordinates {(1,1.01) (2,.66) (3,.58) (4,.32) (5,.05)};
\legend{Human,LLM judge}
\end{axis}
\end{tikzpicture}
\caption{Human and LLM-judge composite scores.}
\label{fig:humanjudgeconcept}
\end{subfigure}%
\hfill
\begin{subfigure}[t]{0.46\textwidth}
\centering
\vspace{0pt}
\begin{tikzpicture}
\begin{axis}[ybar stacked,bar width=23pt,human study axis,
    xticklabel shift=7pt,
    ylabel={Shared instructions (\%)},
    ymin=0,ymax=100,ytick={0,20,40,60,80,100},
    xmin=.5,xmax=4.5,enlarge x limits=false,
    xtick={1,2,3,4},
    xticklabels={{Linear-\\AcT},PCA,CAA,Unsteered},
    legend columns=3,
    nodes near coords={\pgfmathprintnumber[fixed,precision=0]
        {\pgfplotspointmeta}},
    every node near coord/.append style={font=\scriptsize,inner sep=1pt,fill=white,fill opacity=.88,text opacity=1}
]
\addplot[pattern=north east lines,pattern color=wincolor,draw=gray,line width=.35pt,
    every node near coord/.append style={text=black}]
    coordinates {(1,57.89) (2,63.79) (3,57.89) (4,67.24)};
\addplot[pattern=north east lines,pattern color=tiecolor,draw=gray,line width=.35pt,
    every node near coord/.append style={text=textDark}]
    coordinates {(1,21.05) (2,20.69) (3,31.58) (4,27.59)};
\addplot[pattern=north east lines,pattern color=losecolor,draw=gray,line width=.35pt,
    every node near coord/.append style={text=black}]
    coordinates {(1,21.05) (2,15.52) (3,10.53) (4,5.17)};
\legend{Win,Tie,Lose}
\end{axis}
\end{tikzpicture}
\caption{CCS versus each baseline.}
\label{fig:winrate}
\end{subfigure}
\endgroup
\caption{\textbf{Human evaluation on AxBench.}
(a) Human and LLM-judge scores, computed as the harmonic mean of concept,
fluency, and instruction-relevance ratings.
(b) Pairwise wins, ties, and losses for CCS on the human-rated composite
score, against key baselines.}
\label{fig:human-study-results}
\end{figure*}
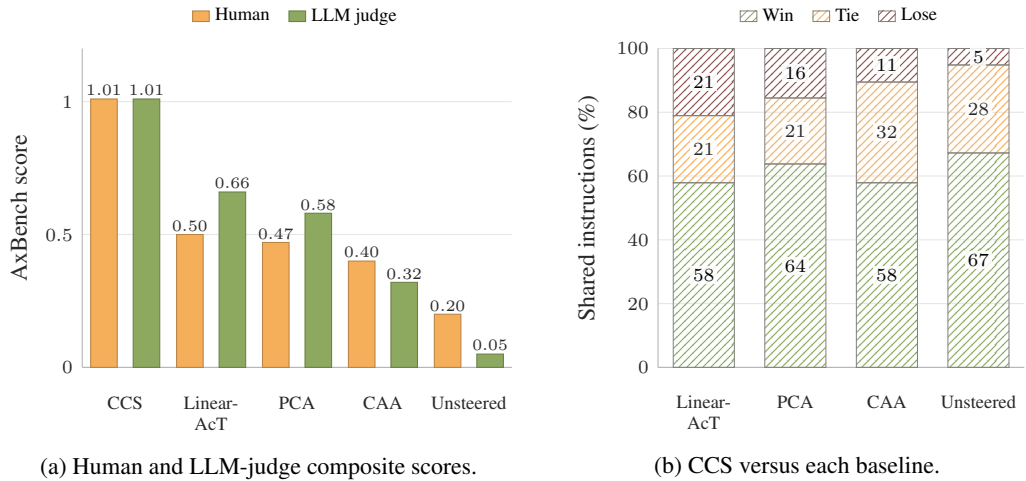

\subsection{Human Eval Study}

To validate our evaluation pipeline, we conduct a human study comparing human ratings against the automated LLM judge across all three AxBench axes. We evaluate $296$ generated responses spanning $10$ held-out concepts and five methods (\textsc{CCS}, \textsc{Linear-AcT}, \textsc{PCA}, \textsc{CAA}, \textsc{Unsteered}) with $23$ human raters who score byte-identical text under identical $0$--$2$ rating anchors (\Cref{app:human_study}).

As shown in~\Cref{fig:humanjudgeconcept}, human ratings and the LLM judge exhibit strong aggregate alignment: specifically, human evaluations reproduce the ranking of all five methods established by the automated judge. On the primary composite metric (AxBench harmonic mean score), \textsc{CCS} achieves the highest score under both human raters ($1.01$) and the LLM judge ($1.01$), leading the next-best method by a wide margin ($0.50$ vs.\ $0.66$ for \textsc{Linear-AcT}). Pairwise win-rate comparisons confirm this dominance (\Cref{fig:winrate}): \textsc{CCS} wins a majority of shared instructions against every baseline ($58$--$67\%$).

More importantly, this gain in concept expression incurs no severe quality cost: human raters judge \textsc{CCS} responses as fluent and on-topic at rates comparable to other steering methods. While absolute ratings differ systematically on individual axes due to judge calibration (the LLM judge applies a stricter standard for fluency than human raters), the relative preferences and overall method rankings are preserved (see~\Cref{app:human_study} for correlation, inter-annotator reliability, and calibration analyses).

\subsection{Ablations}
\label{sec:ablations}

We examine the contributions of CCS's main components for steering performance. First, we evaluate the individual and combined effects of mean-centering and subspace projection in~\Cref{sec:ablation-projection}. We then investigate how performance changes with the number of classes included in the nuisance scatter and grand-mean computation in~\Cref{sec:nuisance-scale}.

\subsubsection{Ablating the Contrastive Subspace Projection}
\label{sec:ablation-projection}

Is CCS's advantage due to mean centering, the subspace projection, or their combination? We compare CAA's difference of means and grand-mean centering, each with and without projection. Figure~\ref{fig:subspace_projection} shows that filtering improves both: CAA gains 0.28--0.36 points, while the grand-mean displacement gains 0.17--0.19. Full CCS performs best across all three models, indicating that the two components provide complementary benefits.

\definecolor{inkmuted}{gray}{0.45}
\definecolor{gridgray}{gray}{0.90}
\definecolor{axisline}{gray}{0.65}

\definecolor{sagegreen}{HTML}{8FAE5D}   
\definecolor{peach}{HTML}{E8A865}       
\definecolor{dustyrose}{HTML}{9E6B70}   

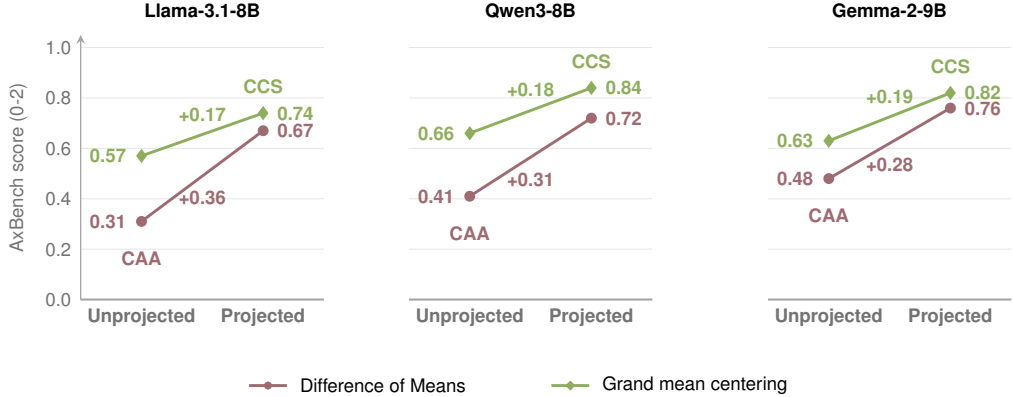
\begin{figure*}[t]
    \centering
    \vspace{-5pt}
    
    \begin{subfigure}[b]{0.32\textwidth}
        \centering
        \begin{tikzpicture}
        \begin{axis}[
            scale only axis,
            width=0.72\textwidth, height=3.5cm,
            title={Llama-3.1-8B},
            title style={font=\sffamily\scriptsize\bfseries, yshift=-3pt},
            ylabel={AxBench score (0-2)},
            label style={font=\sffamily\scriptsize, text=inkmuted},
            ymin=0.0, ymax=1.05,
            ytick={0.0, 0.2, 0.4, 0.6, 0.8, 1.0},
            yticklabels={0.0, 0.2, 0.4, 0.6, 0.8, 1.0},
            yticklabel style={font=\sffamily\scriptsize, text=inkmuted},
            xmin=0.5, xmax=2.5,
            xtick={1, 2},
            xticklabels={Unprojected, Projected},
            xticklabel style={font=\sffamily\scriptsize\bfseries, text=inkmuted},
            tick style={draw=none},
            axis lines*=left,
            x axis line style={axisline, line width=0.8pt},
            y axis line style={->, >=stealth, axisline, line width=0.5pt},
            ymajorgrids=true,
            grid style={gridgray, line width=0.3pt},
            clip=false
        ]
            \addplot[color=sagegreen, line width=1.1pt, mark=diamond*, mark size=1.8pt]
                coordinates {(1, 0.57) (2, 0.74)};
            \addplot[color=dustyrose, line width=1.1pt, mark=*, mark size=1.5pt]
                coordinates {(1, 0.31) (2, 0.67)};

            \node[left, font=\sffamily\scriptsize\bfseries, text=sagegreen] at (axis cs:0.95, 0.57) {0.57};
            \node[above, font=\sffamily\scriptsize\bfseries, text=sagegreen] at (axis cs:2, 0.78) {CCS};
            \node[right, font=\sffamily\scriptsize\bfseries, text=sagegreen] at (axis cs:2.04, 0.74) {0.74};
            \node[above, font=\sffamily\scriptsize\bfseries, text=sagegreen] at (axis cs:1.5, 0.67) {+0.17};

            \node[left, font=\sffamily\scriptsize\bfseries, text=dustyrose] at (axis cs:0.95, 0.31) {0.31};
            \node[below, font=\sffamily\scriptsize\bfseries, text=dustyrose] at (axis cs:1, 0.23) {CAA};
            \node[right, font=\sffamily\scriptsize\bfseries, text=dustyrose] at (axis cs:2.04, 0.67) {0.67};
            \node[below, font=\sffamily\scriptsize\bfseries, text=dustyrose] at (axis cs:1.5, 0.47) {+0.36};
        \end{axis}
        \end{tikzpicture}
    \end{subfigure}\hfill
    \begin{subfigure}[b]{0.32\textwidth}
        \centering
        \begin{tikzpicture}
        \begin{axis}[
            scale only axis,
            width=0.72\textwidth, height=3.5cm,
            title={Qwen3-8B},
            title style={font=\sffamily\scriptsize\bfseries, yshift=-3pt},
            ymin=0.0, ymax=1.05,
            ytick={0.0, 0.2, 0.4, 0.6, 0.8, 1.0},
            yticklabels={},
            xmin=0.5, xmax=2.5,
            xtick={1, 2},
            xticklabels={Unprojected, Projected},
            xticklabel style={font=\sffamily\scriptsize\bfseries, text=inkmuted},
            tick style={draw=none},
            axis lines*=left,
            x axis line style={axisline, line width=0.8pt},
            y axis line style={draw=none},
            ymajorgrids=true,
            grid style={gridgray, line width=0.3pt},
            clip=false
        ]
            \addplot[color=sagegreen, line width=1.1pt, mark=diamond*, mark size=1.8pt]
                coordinates {(1, 0.66) (2, 0.84)};
            \addplot[color=dustyrose, line width=1.1pt, mark=*, mark size=1.5pt]
                coordinates {(1, 0.41) (2, 0.72)};

            \node[left, font=\sffamily\scriptsize\bfseries, text=sagegreen] at (axis cs:0.95, 0.66) {0.66};
            \node[above, font=\sffamily\scriptsize\bfseries, text=sagegreen] at (axis cs:2, 0.88) {CCS};
            \node[right, font=\sffamily\scriptsize\bfseries, text=sagegreen] at (axis cs:2.04, 0.84) {0.84};
            \node[above, font=\sffamily\scriptsize\bfseries, text=sagegreen] at (axis cs:1.5, 0.77) {+0.18};

            \node[left, font=\sffamily\scriptsize\bfseries, text=dustyrose] at (axis cs:0.95, 0.41) {0.41};
            \node[below, font=\sffamily\scriptsize\bfseries, text=dustyrose] at (axis cs:1, 0.33) {CAA};
            \node[right, font=\sffamily\scriptsize\bfseries, text=dustyrose] at (axis cs:2.04, 0.72) {0.72};
            \node[below, font=\sffamily\scriptsize\bfseries, text=dustyrose] at (axis cs:1.5, 0.54) {+0.31};
        \end{axis}
        \end{tikzpicture}
    \end{subfigure}\hfill
    \begin{subfigure}[b]{0.32\textwidth}
        \centering
        \begin{tikzpicture}
        \begin{axis}[
            scale only axis,
            width=0.72\textwidth, height=3.5cm,
            title={Gemma-2-9B},
            title style={font=\sffamily\scriptsize\bfseries, yshift=-3pt},
            ymin=0.0, ymax=1.05,
            ytick={0.0, 0.2, 0.4, 0.6, 0.8, 1.0},
            yticklabels={},
            xmin=0.5, xmax=2.5,
            xtick={1, 2},
            xticklabels={Unprojected, Projected},
            xticklabel style={font=\sffamily\scriptsize\bfseries, text=inkmuted},
            tick style={draw=none},
            axis lines*=left,
            x axis line style={axisline, line width=0.8pt},
            y axis line style={draw=none},
            ymajorgrids=true,
            grid style={gridgray, line width=0.3pt},
            clip=false
        ]
            \addplot[color=sagegreen, line width=1.1pt, mark=diamond*, mark size=1.8pt]
                coordinates {(1, 0.63) (2, 0.82)};
            \addplot[color=dustyrose, line width=1.1pt, mark=*, mark size=1.5pt]
                coordinates {(1, 0.48) (2, 0.76)};

            \node[left, font=\sffamily\scriptsize\bfseries, text=sagegreen] at (axis cs:0.95, 0.63) {0.63};
            \node[above, font=\sffamily\scriptsize\bfseries, text=sagegreen] at (axis cs:2, 0.86) {CCS};
            \node[right, font=\sffamily\scriptsize\bfseries, text=sagegreen] at (axis cs:2.04, 0.82) {0.82};
            \node[above, font=\sffamily\scriptsize\bfseries, text=sagegreen] at (axis cs:1.5, 0.74) {+0.19};

            \node[left, font=\sffamily\scriptsize\bfseries, text=dustyrose] at (axis cs:0.95, 0.48) {0.48};
            \node[below, font=\sffamily\scriptsize\bfseries, text=dustyrose] at (axis cs:1, 0.40) {CAA};
            \node[right, font=\sffamily\scriptsize\bfseries, text=dustyrose] at (axis cs:2.04, 0.76) {0.76};
            \node[below, font=\sffamily\scriptsize\bfseries, text=dustyrose] at (axis cs:1.5, 0.60) {+0.28};
        \end{axis}
        \end{tikzpicture}
    \end{subfigure}

    \vspace{2.5ex}
    
    \begin{tikzpicture}
        \draw[color=dustyrose, line width=1.1pt] (0,0) -- (0.5,0);
        \fill[dustyrose] (0.25,0) circle (1.5pt);
        \node[right, font=\sffamily\scriptsize, text=black] at (0.55,0) {Difference of Means};

        \draw[color=sagegreen, line width=1.1pt] (4.0,0) -- (4.5,0);
        \node[diamond, fill=sagegreen, inner sep=0pt, minimum size=4pt] at (4.25,0) {};
        \node[right, font=\sffamily\scriptsize, text=black] at (4.55,0) {Grand mean centering};
    \end{tikzpicture}
    
    \caption{\textbf{Effect of subspace projection on AxBench.} Projection improves both directions in all three models. y-axis displacement shows scale of improvement. Grand-mean centered directions consistently outperform the naive difference of means, while projection is effective on both baselines.}
    \vspace{-10pt}
    \label{fig:subspace_projection}
\end{figure*}

\subsubsection{Robustness to the nuisance-pool size}
\label{sec:nuisance-scale}

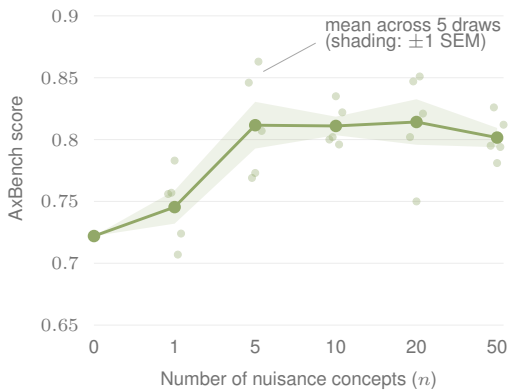
\begin{wrapfigure}[18]{l}{0.48\textwidth}
\centering
\vspace{-10pt}
\definecolor{axbgreen}{HTML}{92A868}

\begin{tikzpicture}
\begin{axis}[
    width=7.0cm, height=6.0cm,
    xlabel={Number of nuisance concepts ($n$)},
    ylabel={AxBench score},
    label style={font=\sffamily\scriptsize, text=inkmuted},
    xtick={1,2,3,4,5,6},
    xticklabels={0,1,5,10,20,50},
    tick label style={font=\sffamily\scriptsize, text=inkmuted},
    ymin=0.65, ymax=0.92,
    ytick={0.65,0.70,0.75,0.80,0.85,0.90},
    axis lines=left,
    axis line style={draw=none},
    tick style={draw=none},
    ymajorgrids=true,
    grid style={gray!15, line width=0.2pt},
    clip=false,
]

\addplot[name path=upper, draw=none, forget plot]
    coordinates {(1,0.7220) (2,0.7588) (3,0.8305) (4,0.8185) (5,0.8326) (6,0.8094)};
\addplot[name path=lower, draw=none, forget plot]
    coordinates {(1,0.7220) (2,0.7320) (3,0.7927) (4,0.8035) (5,0.7958) (6,0.7938)};
\addplot[axbgreen, opacity=0.15, forget plot] fill between[of=upper and lower];

\addplot[only marks, mark=*, mark size=1.3pt, axbgreen, opacity=0.35, forget plot]
    coordinates {(1.92,0.756) (1.96,0.757) (2,0.783) (2.04,0.707) (2.08,0.724)};
\addplot[only marks, mark=*, mark size=1.3pt, axbgreen, opacity=0.35, forget plot]
    coordinates {(2.92,0.846) (2.96,0.769) (3,0.773) (3.04,0.863) (3.08,0.807)};
\addplot[only marks, mark=*, mark size=1.3pt, axbgreen, opacity=0.35, forget plot]
    coordinates {(3.92,0.800) (3.96,0.802) (4,0.835) (4.04,0.796) (4.08,0.822)};
\addplot[only marks, mark=*, mark size=1.3pt, axbgreen, opacity=0.35, forget plot]
    coordinates {(4.92,0.802) (4.96,0.847) (5,0.750) (5.04,0.851) (5.08,0.821)};
\addplot[only marks, mark=*, mark size=1.3pt, axbgreen, opacity=0.35, forget plot]
    coordinates {(5.92,0.795) (5.96,0.826) (6,0.781) (6.04,0.794) (6.08,0.812)};

\addplot[color=axbgreen, mark=*, mark size=1.8pt, line width=1.1pt, forget plot]
    coordinates {(1,0.7220) (2,0.7454) (3,0.8116) (4,0.8110) (5,0.8142) (6,0.8016)};

\draw[gray!70, line width=0.3pt] (axis cs:3.1,0.855) -- (axis cs:3.75,0.878);
\node[align=left, font=\sffamily\scriptsize, text=inkmuted, anchor=west]
    at (axis cs:3.75,0.885) {mean across 5 draws\\[-2pt](shading: $\pm$1 SEM)};

\end{axis}
\end{tikzpicture}
\caption{\textbf{Mean AxBench scores for $N$ nuisance concepts.} Gray dots indicate individual draws.}
\label{fig:nuisance-pool-size}
\end{wrapfigure}

To evaluate how nuisance pool size affects CCS (\Cref{sec:method}), we fix a 20-concept evaluation subset from AxBench and sweep $N \in \{0, 1, 5, 10, 20, 50\}$. Nuisance concepts are sampled uniformly from a disjoint 470-concept pool ($N=0$ includes only the base negative class). For each $N > 0$, we report the mean and standard error across 5 independent random draws. Hyperparameters are fixed at $k=16$, with $r \in \{0.3, 1, 3, 10, 30, 50, 100\}$ selected per concept.

As shown in \Cref{fig:nuisance-pool-size}, adding just 5--10 nuisance concepts yields a marked improvement over the $N=0$ baseline. Performance peaks at $N=20$ (0.814 vs.\ 0.811 at $N=10$) before declining at $N=50$. This confirms that a small nuisance pool (5--10 concepts) captures most benefits of contrastive centering without introducing excessive noise or correlated concepts.

\subsection{Steering Multi-Turn Conversations}
\label{sec:multiturn}

We extend our evaluation to long-horizon steering stability over $20$-turn conversations using \textsc{Qwen3-8B} across five held-out AxBench concepts (details in~\Cref{app:multiturn_details}). Turn~$0$ reproduces standard single-turn steering; turns~$1$--$19$ apply a fixed sequence of subject-agnostic, anaphoric follow-ups requiring context retention. We benchmark \textsc{CCS} against additive (CAA, RepE, PCA), projection (SEA, Linear-AcT), unsteered, and system-prompting baselines. 
All steering methods select hyperparameters via 3-fold cross-validation at turn 0.
\begin{wrapfigure}[17]{r}{0.52\textwidth}
\centering
\input{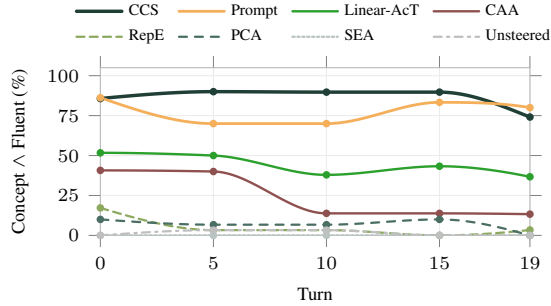}
\vspace{-10pt}
\caption{\textbf{Multi-turn steering}: Fraction of replies that are simultaneously on-concept ($\geq1$) and fluent ($\geq1$) over 20 turns. Curves are interpolated 5-turn intervals.}
\label{fig:mt-perturn}
\vspace{-9pt}
\end{wrapfigure}
Because a steered reply is only useful if it retains both topic relevance and output quality, our primary metric is the joint fraction of replies satisfying both concept ($\ge 1$) and fluency ($\ge 1$) ratings from an LLM judge (truncating incomplete trailing sentences at the $300$-token budget).

\Cref{fig:mt-perturn} (interpolated between turns 0, 5, 10, 15 and 19; see~\Cref{tab:mt} in~\Cref{app:multiturn_details}) shows that \textsc{CCS} is the most robust method, sustaining a $93\% \!\rightarrow\! 80\%$ joint concept-fluency score over 20 turns while its on-concept score rises ($95\% \!\rightarrow\! 98\%$). Other steering baselines degrade under persistent KV caching: \textsc{CAA} collapses ($36\% \!\rightarrow\! 10\%$) due to repetition loops, \textsc{RepE} drops to $2\%$, \textsc{PCA} remains below $9\%$, and \textsc{SEA} rarely expresses concepts ($\le 2\%$). \textsc{Linear-AcT} is the strongest activation-steering baseline, retaining concept expression across turns ($52\% \!\rightarrow\! 37\%$ joint) though still well below \textsc{CCS}. System prompting preserves fluency but suffers prompt drift ($91\% \!\rightarrow\! 71\%$), allowing \textsc{CCS} to overtake it by turn 20.

\section{Limitations and Societal Impact}
CCS operates as a static, single-layer residual-stream intervention; extending contrastive projection across multiple layers or attention heads \citep{li2023inference} could enable finer-grained control. Additionally, accurately estimating scatter matrices $\mathbf{S}_c$ requires sufficient sample activations per concept, which can become ill-conditioned in extreme low-data regimes. Finally, while CCS improves multi-turn stability (\Cref{sec:multiturn}), static steering over extended horizons ($>20$ turns) may still benefit from dynamic decay schedules.

Activation steering is dual-use. On one hand, CCS enables lightweight, training-free safety alignment, such as mitigating sycophancy and hallucinations, without costly retraining. On the other hand, latent vector interventions can be exploited to bypass safety guardrails or suppress refusal behavior \citep{arditi2024refusal}. Clarifying how second-order geometry isolates concept directions advances both effective model control and defenses against inference-time manipulation.

\section{Conclusion}
In this work, we introduced Contrastive Covariance Steering (CCS), a training-free method that automatically identifies and removes confound variation like incidental topic or style that ordinary steering directions inadvertently pick up alongside the target concept, by constructing a low-dimensional subspace that maximizes target concept variability, minimizing variability in other concepts.
We demonstrate, by testing on four major benchmarks and three open-weight LLMs, how our method consistently outperforms other closed-form methods for finding LLM steering directions. Ablations show that both our grand mean centering and subspace projection independently and jointly increase steering vector quality. A controlled diagnostic on AxBench where we explicitly inject confounds by oversampling
further justifies our method's variance objective. Finally, we demonstrate how the directions obtained by our method lead to significantly less degradation during long multi-turn generation.

\section*{AI use statement}

In this work, we used generative AI tools to:

\begin{itemize}
\item Generate evaluation prompts for the Emotion benchmark \cite{sofroniew2026emotion} (see~\Cref{app:eval-prompts}).
\item Generate the multi-turn evaluation questions in~\Cref{sec:multiturn}.
\item Minor polishing/drafting certain written passages.
\end{itemize}

We have \textbf{not} used generative
AI tools to develop theoretical models or conceptual frameworks, to
formulate or prove mathematical claims.

We have thoroughly reviewed all AI-assisted work. The synthetic data was inspected by
all authors for label correctness and for the absence of unintended
confounds before use.
All the contents of the paper that were drafted using AI tools have been re-written by the authors. We take responsibility for the final content of this work, including
text, claims or artifacts produced with the aid of generative AI.




\section*{Ethics Statement}

CCS can steer concepts beyond the emotions and personality traits
studied here. This creates a dual-use risk: improvements in steering
could also make it easier to elicit deception, sycophancy, or other
harmful behaviors. These risks already exist with simpler methods
such as difference-of-means steering, but stronger control may
amplify them. CCS requires access to internal activations, so it
cannot be applied directly through standard text-only APIs.
Its learned directions may also support defensive applications,
such as monitoring undesirable traits, although their reliability
for this purpose requires separate evaluation.

LLM judges can reproduce biases in their training data, and human evaluation provides only a partial check on their judgments. The personality and emotion
taxonomies we use reflect particular theoretical and cultural traditions rather than universal descriptions of human behavior. 

\section*{Reproducibility Statement}

All components needed to reproduce our results are released or specified.
The steering-vector construction is given in closed form in~\Cref{sec:method} (\Crefrange{eq:grand_mean}{eq:steering_vector}, with the injection rule of~\Cref{eq:per_token_norm}),
requires no gradient-based training, and depends only on a layer index and
the two hyperparameters $(k,r)$. Information on baseline implementation is given in~\Cref{sec:experimental-setup} and~\Cref{app:hparams-selection}.

All the code and data needed to recreate the method and evaluations will be made public after acceptance.

\bibliography{iclr2026_conference}
\bibliographystyle{iclr2026_conference}

\clearpage
\appendix

\section{Implementation Details}
\label{app:implementation_details}

\subsection{Generation configuration}
\label{app:generation-config}

All steered and unsteered completions use greedy decoding
(\texttt{do\_sample=False}), a repetition penalty of 1.0, and a maximum of 300
new tokens, with each model's native dtype (\texttt{torch\_dtype="auto"},
bfloat16 in practice for all three models) and left-padded batched inputs.
Prompts are formatted with each model's own chat template
(\texttt{apply\_chat\_template}, a single user turn, generation prompt
appended); for Qwen3-8B we disable its extended thinking mode
(\texttt{enable\_thinking=False}) so its output is a direct response rather
than a reasoning trace. Attention is computed with FlashAttention-2 where a
compatible wheel is available, falling back to PyTorch's native
scaled-dot-product attention otherwise; we did not observe this choice to
affect results at the aggregate level.

\subsection{Evaluation prompts}
\label{app:eval-prompts}

AxBench is evaluated on its 50 held-out concepts with the benchmark's own 10
AlpacaEval instructions per concept, and BIG5 on the NPTI prompt set (459
prompts, 89--100 per trait). For the Emotion benchmark we generated the
evaluation prompts ourselves with a generative AI tool: 55 short,
emotion-neutral writing tasks about everyday activities (e.g.\ ``Write a short
story about someone going grocery shopping.''), obtained by expanding 5 seed
prompts to 55. The same 55 prompts are used for all seven target emotions, so
every Emotion operating point is judged on $7\times55=385$ generations. The
prompts were inspected by the authors before use (see the AI use statement).

\subsection{Steering strength and per-token normalization}
\label{app:per-token-norm}

The steering strength $\alpha$ of~\Cref{eq:normalized_steering} is not a
single constant shared across layers, models and token positions. Residual-stream
norms vary considerably with depth and between model families, so one fixed
$\alpha$ would amount to a negligible edit at one injection site and an
overwhelming one at another, and would not transfer between the three models we
evaluate. We therefore sweep a dimensionless \emph{relative} strength
$\alpha$:
\begin{equation}
    \mathbf{h}_{\ell,s}^{\,\mathrm{steer}}
    =\mathbf{h}_{\ell,s}
    +\alpha\,\lVert\mathbf{h}_{\ell,s}\rVert_2\,\widehat{\mathbf{v}}_t .
    \label{eq:per_token_norm}
\end{equation}
Thus $\alpha$ is the magnitude of the injected component expressed as a fraction
of the current activation's own norm, and it carries the same meaning at every
layer and in every model. 
CCS, CAA, RepE and PCA share this identical injection
code and differ only in the vector $\mathbf{v}_t$; for each of them $\alpha$ is
swept over $[0,1]$ in steps of $0.1$. The remaining baselines are not additive
and therefore have no directly comparable knob: Linear-AcT and SEA interpolate
between the unedited activation and a transported or projected version of it,
and Angular Steering rotates the activation inside a fixed two-dimensional
plane, so their sweeps range over an interpolation weight and a rotation angle
respectively.

\subsection{LLM judge instructions}
\label{app:judge-prompts}

Every domain except TruthfulQA is scored by an LLM judge
(\texttt{Qwen2.5-32B-Instruct}); TruthfulQA uses the two fine-tuned judges released with
the benchmark (\texttt{allenai/truthfulqa-\{truth,info\}-judge-llama2-7B} in Huggingface hub). The exact instructions are reproduced below, with \texttt{\{braces\}}
marking the fields substituted per item.

\paragraph{AxBench.} Each generation is rated on three independent axes; the reported score is the harmonic mean of the three ratings, following \citet{wu2025axbench}. Each axis is a separate judge call, so no rating can condition on another.

\textit{Concept relevance.}
\begin{quote}\small\ttfamily\noindent
[System]\\
Please act as an impartial judge and evaluate whether the specified concept is presented in the sentence fragment provided below. Focus solely on whether the concept is clearly incorporated, without regard for grammar or logical coherence.\\
\medskip
Begin your evaluation by providing a short explanation of whether the concept is clearly incorporated. Point out references to the concept in the fragment, noting any minimal or awkward incorporations. After providing your explanation, rate the concept's relevance on a scale from 0 to 2, where 0 indicates the concept is not present at all, 1 indicates the concept is somewhat present but minimally or awkwardly incorporated, and 2 indicates the concept is more fully and effectively incorporated, with stronger and more natural integration. Provide your rating using this exact format: "Rating: [[score]]".\\
\medskip
[Concept Start]\\
\{concept\}\\
\texttt{[Concept End]}\\
\medskip
[Sentence Fragment Start]\\
\{sentence\}\\
\texttt{[Sentence Fragment End]}
\end{quote}

\textit{Instruction relevance.}
\begin{quote}\small\ttfamily\noindent
[System]\\
Please act as an impartial judge and evaluate whether the sentence fragment provided below is related to the instruction. Focus solely on the degree of relatedness in terms of topic, regardless of grammar, coherence, or informativeness.\\
\medskip
Begin your evaluation by providing a brief explanation of whether the sentence is related to the instruction, and point out references related to the instruction. After providing your explanation, rate the instruction relevance on a scale from 0 to 2, where 0 indicates the sentence is unrelated to the instruction, 1 indicates it is somewhat related but only minimally or indirectly relevant in terms of topic, and 2 indicates it is more clearly and directly related to the instruction. Provide your rating using this exact format: "Rating: [[score]]".\\
\medskip
[Instruction Start]\\
\{instruction\}\\
\texttt{[Instruction End]}\\
\medskip
[Sentence Fragment Start]\\
\{sentence\}\\
\texttt{[Sentence Fragment End]}
\end{quote}

\textit{Fluency.}
\begin{quote}\small\ttfamily\noindent
[System]\\
Please act as an impartial judge and evaluate the fluency of the sentence fragment provided below. Focus solely on fluency, disregarding its completeness, relevance, coherence with any broader context, or informativeness.\\
\medskip
Begin your evaluation by briefly describing the fluency of the sentence, noting any unnatural phrasing, awkward transitions, grammatical errors, or repetitive structures that may hinder readability. After providing your explanation, rate the sentence's fluency on a scale from 0 to 2, where 0 indicates the sentence is not fluent and highly unnatural (e.g., incomprehensible or repetitive), 1 indicates it is somewhat fluent but contains noticeable errors or awkward phrasing, and 2 indicates the sentence is fluent and almost perfect. Provide your rating using this exact format: "Rating: [[score]]".\\
\medskip
[Sentence Fragment Start]\\
\{sentence\}\\
\texttt{[Sentence Fragment End]}
\end{quote}

\paragraph{BIG5 and Emotion.} Both use the same instructions, with \texttt{\{construct\}} set to \texttt{personality trait} for BIG5 and \texttt{emotion} for Emotion, and \texttt{\{emotion\}} naming the target trait or emotion. 

\begin{quote}\small\ttfamily\noindent
You are an expert annotator rating how strongly a piece of text expresses a specific \{construct\}.\\
\medskip
The text was generated in response to this WRITING TASK:\\
"""\{task\}"""\\
\medskip
TEXT TO RATE:\\
"""\{text\}"""\\
\medskip
Rate the intensity with which the TEXT expresses this \{construct\}, on a 0-100 scale:\\
- \{emotion\}\\
\medskip
Scale anchors: 0 = absent, 25 = faint, 50 = clearly present, 75 = strong, 100 = unmistakable / dominant.\\
\medskip
Also rate two SEPARATE quality dimensions (do not blend them together, and do not let them affect the \{construct\} rating above):\\
- "fluency": 0-100 --- purely the well-formedness of the prose: grammar, syntax, no run-ons / repetition / word-salad. IGNORE both the \{construct\}-related content AND whether it answers the task. A vivid, off-topic, or exaggerated passage can still be perfectly fluent (high score) as long as the sentences are grammatical and read naturally.\\
  Anchors: 0 = word-salad / garbled | 25 = mostly broken, hard to follow | 50 = understandable but awkward or repetitive | 75 = fluent with minor issues | 100 = perfectly natural, fluent prose.\\
  Calibration examples (judge ONLY the prose mechanics here, ignore their content):\\
    fluency=15: "the the market was was going to buy the buy things she she needed needed for for the day and the person walking walking to the the store and buying the the milk and milk was was there and buying and buying more more things for the the house today today was was good good shopping trip trip."\\
    fluency=55: "She went to the store to buy some the milk and bread today. It was necessary because the fridge was empty of food things. She walked there and bought the items and then she carried them back to home in bags that she has brought with her from the house before going."\\
    fluency=95: "She grabbed her canvas bag and headed to the corner store, the autumn air crisp against her skin. Inside, she moved briskly between the aisles, picking out milk, bread, and a few ripe tomatoes before heading home."\\
- "prompt\_adherence": 0-100 --- how well the text stays ON the WRITING TASK above (right format, topic, and intent). IGNORE fluency and \{construct\} here; judge only on-task-ness.\\
  Anchors: 0 = ignores the task entirely / different topic | 50 = loosely related but drifts off | 100 = fully addresses the task as asked.\\
  IMPORTANT: a text that performs the task while strongly expressing the target \{construct\} is still fully on-task --- \{construct\} expression is NOT drift. Only penalize this score for actual topic, format, or intent drift (e.g. abandoning the described scenario/activity, switching to an unrelated subject, ignoring the requested format). A strongly \{construct\}-expressive version of the same task is not, by itself, evidence of lower adherence.\\
\medskip
Return ONLY a JSON object, no preamble:\\
\{\{"effectiveness": <0-100>, "fluency": <0-100>, "prompt\_adherence": <0-100>\}\}
\end{quote}

\paragraph{TruthfulQA.} Scored not by a prompted judge but by the two fine-tuned classifiers released with the benchmark, \texttt{allenai/truthfulqa-truth-judge-llama2-7B} and \texttt{-info-judge-llama2-7B}. Each greedily generates a single token, read as yes/no; we report $\mathrm{True}\% \times \mathrm{Info}\%/100$. Their input formats, taken from the model cards, are:

\begin{quote}\small\ttfamily\noindent
Truthfulness judge:\\
  Q: \{question\}\\
  A: \{answer\}\\
  True:\\
\medskip
Informativeness judge:\\
  Q: \{question\}\\
  A: \{answer\}\\
  Helpful:
\end{quote}

Because these judges are fine-tuned rather than prompted, TruthfulQA scores are not comparable to the 0--100 or 0--2 scales used elsewhere, and are not affected by the choice of judge model.

\section{Human Study Protocol}
\label{app:human_study}

\paragraph{Concepts Used:}
\emph{textures and texture-related properties in a technical context}; \emph{words related to
fishing activities and experiences}; \emph{phrases that indicate accolades or recognition};
\emph{negative critiques or observations related to the quality or authenticity of artistic
expressions}; \emph{phrases related to dog training and behavior management}; \emph{terms
associated with medical or scientific specifics}; \emph{terms and phrases related to regulatory
compliance and usage of pharmaceuticals}; \emph{references to temporal states and experiences
related to aging}; \emph{references to islands and geographical locations}; and \emph{terms
related to product development and quality assessment}. 

We sampled $10$ of the $35$ concepts in the held-out AxBench-50 pool, excluding code and math concepts so that generations are human-readable.
We also exclude
function-word and punctuation concepts (e.g. \emph{punctuation marks and their occurrences}), where successful steering is indistinguishable
from degraded generation. Each concept contributes $6$
AlpacaEval evaluation prompts, giving $60$ instructions $\times$ $5$ methods $= 300$ candidate
items, of which $296$ enter the study ($4$ dropped for judge-API failures). Every method used its best selected hyperparameters from the main experiments in~\Cref{sec:main_experiment}. Responses are shown exactly as generated,
with no display-side truncation, so raters and the LLM judge score byte-identical text; the
$128$-token generation cap means a response sometimes stops mid-sentence, which the fluency
instructions explicitly instructs raters to disregard.

\paragraph{Rating interface.}
Each item is rated on three screens in a fixed order with progressive reveal and no
back-navigation: fluency (response alone), instruction-following (response $+$ instruction), then
concept (response $+$ concept). This mirrors the information sets of AxBench's three judge
prompts---its fluency judge sees neither the instruction nor the concept, its instruction judge
sees the instruction, its concept judge sees the concept---and prevents each judgement from
contaminating the next. Placeholders in braces are substituted per item.

\paragraph{Screen 1 --- Fluency.}
\begin{quote}\small
\textbf{Response:} \texttt{\{response\}}

\medskip
\textbf{Rate how well-written this is --- the quality of the English only.}

\medskip
Ignore everything except the writing:
\begin{itemize}\itemsep0pt
\item Ignore whether the content is true, useful, or on topic. Text can be completely off topic
and still be perfectly well written.
\item \textbf{Ignore formatting.} Markdown, bullet points, numbered lists, headings, bold text,
tables and code blocks are normal ways to write an answer. Do not lower the rating for them, and
do not raise it for them.
\item Ignore whether it feels complete.
\end{itemize}

\textbf{Do lower the rating for:} broken grammar, sentences that don't make sense, phrases or
sentences repeated over and over, abrupt shifts that make it hard to follow, garbled or invented
words.

\medskip
\begin{itemize}\itemsep0pt
\item[] $\bigcirc$\; \textbf{0} --- Not fluent and highly unnatural: incomprehensible, or repetitive.
\item[] $\bigcirc$\; \textbf{1} --- Somewhat fluent, but contains noticeable errors or awkward phrasing.
\item[] $\bigcirc$\; \textbf{2} --- Fluent and almost perfect.
\end{itemize}
\end{quote}

\paragraph{Screen 2 --- Instruction-following.}
\begin{quote}\small
\textbf{The user asked:} \texttt{\{instruction\}}

\medskip
\textbf{The response was:} \texttt{\{response\}}

\medskip
\textbf{Rate how related the response is to the request.}

\medskip
Judge \textbf{only} relatedness of topic. Ignore grammar, writing quality and formatting. Ignore
whether the answer is correct, complete or useful --- only whether it is \emph{about} what was
asked.

\medskip
\begin{itemize}\itemsep0pt
\item[] $\bigcirc$\; \textbf{0} --- The response is unrelated to the request.
\item[] $\bigcirc$\; \textbf{1} --- Somewhat related, but only minimally or indirectly relevant in topic.
\item[] $\bigcirc$\; \textbf{2} --- More clearly and directly related to the request.
\end{itemize}
\end{quote}

\paragraph{Screen 3 --- Concept.}
\begin{quote}\small
\textbf{Concept:} \texttt{\{concept\_text\}}

\medskip
\textbf{The response was:} \texttt{\{response\}}

\medskip
\textbf{How much does this concept show up in the response?}

\medskip
You are looking for one thing only: whether this concept is present in the text, and how fully.
\begin{itemize}\itemsep0pt
\item \textbf{Ignore the writing quality.} Grammar, repetition and awkward phrasing are rated
separately and should not affect this score.
\item \textbf{Ignore formatting.} A bulleted list, a numbered recipe, a heading or an
assistant-style answer is a normal way to write. A concept woven into a list counts exactly as
much as the same concept woven into a paragraph.
\item \textbf{Ignore whether the concept belongs there.} The concept may have been deliberately
pushed into a response about something else entirely. That is expected. Text that brings the
concept into an unrelated topic still scores for presence.
\end{itemize}

\medskip
\begin{itemize}\itemsep0pt
\item[] $\bigcirc$\; \textbf{0} --- The concept is not present at all.
\item[] $\bigcirc$\; \textbf{1} --- The concept is somewhat present, but only minimally or
awkwardly worked in --- a passing mention, a single forced word.
\item[] $\bigcirc$\; \textbf{2} --- The concept is fully and effectively incorporated, woven
through the text rather than tacked on.
\end{itemize}
\end{quote}

\paragraph{Raters, onboarding, and sample distribution.}
Twenty-three raters each completed $17$--$18$ items in approximately $30$ minutes, blind to method, with randomised item order and no rater seeing two responses to the same instruction. Onboarding presented the three prompts followed by two gold items drawn from the real pool. The first is a generation that repeats the same sentence verbatim three times and must receive fluency $=0$; the second is an unsteered response to an unrelated translation request, rated against the concept \emph{terms related to product development and quality assessment}, and must receive concept $=0$. Raters failing either gold item were excluded according to a pre-registered rule.

The study evaluated $296$ total responses ($60$ instructions $\times$ $5$ methods, minus $4$ dropped for judge-API failures). Annotator allocation averaged $1.39$ raters per sample: $202$ samples were evaluated by $1$ rater, $75$ by $2$, $18$ by $3$, and $1$ by $4$ raters. 

\paragraph{Inter-annotator agreement.}
A subset of $94$ unique items were multi-rated ($208$ total ratings) to evaluate inter-rater reliability. As shown in~\Cref{tab:inter_annotator}, exact agreement ranges from $54.8\%$ to $71.1\%$. Because fluency ratings are heavily skewed toward high quality ($77\%$ rated at ceiling), traditional chance-corrected metrics can severely penalize marginal imbalances; we therefore report Gwet's $\mathrm{AC2}$~\citep{gwet2014handbook}, demonstrating high inter-rater agreement ($\mathrm{AC2} = 0.780$--$0.796$) on fluency and adherence.

\begin{table}[h]
\centering
\small
\caption{Inter-annotator agreement across $94$ multi-rated items ($208$ total ratings).}
\label{tab:inter_annotator}
\begin{tabular}{lcc}
\toprule
\textbf{Axis} & \textbf{Gwet's $\mathrm{AC2}$} & \textbf{Exact Agreement} \\
\midrule
Fluency   & $0.780$ & $71.1\%$ \\
Adherence & $0.796$ & $67.4\%$ \\
Concept   & $0.508$ & $54.8\%$ \\
\bottomrule
\end{tabular}
\end{table}

\paragraph{Human--judge agreement and calibration shift.}
Item-level correlation between human and LLM judge ratings yields Spearman $\rho = 0.548$ for concept presence, $\rho = 0.548$  for instruction adherence, and $\rho = 0.194$ for fluency. The lower correlation on fluency reflects a systematic calibration shift between human raters and the automated judge rather than a disagreement regarding relative model quality. Human raters adopt a more lenient standard (reading between the lines and ignoring minor mid-sentence truncations), placing $228/296$ items ($77.0\%$) at the maximum score of $2$. In contrast, the LLM judge enforces a stricter threshold, assigning $175/296$ items a score of $1$. On average, human raters score fluency $0.33$ points higher than the LLM judge. This strong ceiling effect in human ratings ($77\%$ at score $2$) severely restricts range variance, which depresses correlation metrics ($\rho$) while maintaining strong nominal agreement on overall output acceptability.

\begin{figure*}[t]
    \centering
    \begin{subfigure}[t]{0.32\textwidth}
        \centering
        \begin{tikzpicture}
        \begin{axis}[
            ybar stacked,
            width=\textwidth,
            height=5.2cm,
            bar width=16pt,
            enlarge x limits=0.18,
            ylabel={Percentage of shared instructions},
            ylabel shift=6pt,
            yticklabel style={inner sep=1pt, font=\footnotesize},
            symbolic x coords={Linear-AcT, PCA, CAA, Unsteered},
            xtick=data,
            ymin=0,
            ymax=100,
            ytick={0,20,40,60,80,100},
            xticklabel style={font=\scriptsize, rotate=30, anchor=north east},
            nodes near coords={\pgfmathprintnumber[fixed, precision=0]{\pgfplotspointmeta}},
            every node near coord/.append style={font=\scriptsize, color=black},
            major grid style={gray!30},
            grid=major,
        ]
        \addplot+[
            ybar,
            pattern=north east lines,
            pattern color=wincolor,
            draw=gray
        ] coordinates {
            (Linear-AcT,29.82)
            (PCA,22.41)
            (CAA,14.04)
            (Unsteered,8.62)
        };
        \addplot+[
            ybar,
            pattern=north east lines,
            pattern color=tiecolor,
            draw=gray
        ] coordinates {
            (Linear-AcT,49.12)
            (PCA,51.72)
            (CAA,64.91)
            (Unsteered,58.62)
        };
        \addplot+[
            ybar,
            pattern=north east lines,
            pattern color=losecolor,
            draw=gray
        ] coordinates {
            (Linear-AcT,21.05)
            (PCA,25.86)
            (CAA,21.05)
            (Unsteered,32.76)
        };
        \end{axis}
        \end{tikzpicture}
        \caption{Fluency}
        \label{fig:winrate-fluency}
    \end{subfigure}
    \hfill
    \begin{subfigure}[t]{0.32\textwidth}
        \centering
        \begin{tikzpicture}
        \begin{axis}[
            ybar stacked,
            width=\textwidth,
            height=5.2cm,
            bar width=16pt,
            enlarge x limits=0.18,
            ylabel={\hphantom{Percentage of shared instructions}},
            ylabel shift=6pt,
            yticklabel style={inner sep=1pt, font=\footnotesize},
            symbolic x coords={Linear-AcT, PCA, CAA, Unsteered},
            xtick=data,
            ymin=0,
            ymax=100,
            ytick={0,20,40,60,80,100},
            xticklabel style={font=\scriptsize, rotate=30, anchor=north east},
            nodes near coords={\pgfmathprintnumber[fixed, precision=0]{\pgfplotspointmeta}},
            every node near coord/.append style={font=\scriptsize, color=black},
            major grid style={gray!30},
            grid=major,
        ]
        \addplot+[
            ybar,
            pattern=north east lines,
            pattern color=wincolor,
            draw=gray
        ] coordinates {
            (Linear-AcT,45.61)
            (PCA,36.21)
            (CAA,21.05)
            (Unsteered,8.62)
        };
        \addplot+[
            ybar,
            pattern=north east lines,
            pattern color=tiecolor,
            draw=gray
        ] coordinates {
            (Linear-AcT,35.09)
            (PCA,36.21)
            (CAA,45.61)
            (Unsteered,50.00)
        };
        \addplot+[
            ybar,
            pattern=north east lines,
            pattern color=losecolor,
            draw=gray
        ] coordinates {
            (Linear-AcT,19.30)
            (PCA,27.59)
            (CAA,33.33)
            (Unsteered,41.38)
        };
        \end{axis}
        \end{tikzpicture}
        \caption{Instruction relevance}
        \label{fig:winrate-adherence}
    \end{subfigure}
    \hfill
    \begin{subfigure}[t]{0.32\textwidth}
        \centering
        \begin{tikzpicture}
        \begin{axis}[
            ybar stacked,
            width=\textwidth,
            height=5.2cm,
            bar width=16pt,
            enlarge x limits=0.18,
            ylabel={\hphantom{Percentage of shared instructions}},
            ylabel shift=6pt,
            yticklabel style={inner sep=1pt, font=\footnotesize},
            symbolic x coords={Linear-AcT, PCA, CAA, Unsteered},
            xtick=data,
            ymin=0,
            ymax=100,
            ytick={0,20,40,60,80,100},
            xticklabel style={font=\scriptsize, rotate=30, anchor=north east},
            nodes near coords={\pgfmathprintnumber[fixed, precision=0]{\pgfplotspointmeta}},
            every node near coord/.append style={font=\scriptsize, color=black},
            major grid style={gray!30},
            grid=major,
        ]
        \addplot+[
            ybar,
            pattern=north east lines,
            pattern color=wincolor,
            draw=gray
        ] coordinates {
            (Linear-AcT,47.37)
            (PCA,53.45)
            (CAA,59.65)
            (Unsteered,74.14)
        };
        \addplot+[
            ybar,
            pattern=north east lines,
            pattern color=tiecolor,
            draw=gray
        ] coordinates {
            (Linear-AcT,31.58)
            (PCA,24.14)
            (CAA,28.07)
            (Unsteered,22.41)
        };
        \addplot+[
            ybar,
            pattern=north east lines,
            pattern color=losecolor,
            draw=gray
        ] coordinates {
            (Linear-AcT,21.05)
            (PCA,22.41)
            (CAA,12.28)
            (Unsteered,3.45)
        };
        \end{axis}
        \end{tikzpicture}
        \caption{Concept}
        \label{fig:winrate-concept}
    \end{subfigure}
    \par\vspace{4pt}
    \begin{tikzpicture}
        \foreach \lbl/\col/\x in {Win/wincolor/0, Tie/tiecolor/1.6, Lose/losecolor/3.2} {
            \draw[pattern=north east lines, pattern color=\col, draw=gray]
                (\x,0) rectangle ++(0.32,0.22);
            \node[right, font=\footnotesize, inner sep=2pt] at (\x+0.32,0.11) {\lbl};
        }
    \end{tikzpicture}
    \caption{\textbf{Win rate of \textsc{CCS} per rating axis.} Each panel compares \textsc{CCS} against one competing method on the instructions both arms answered ($n=57$--$58$ per pair, the variation coming from the four items dropped for a NaN judge score), using the item-mean human rating on that axis. Ties are frequent because the scale has three points, and are kept in the denominator rather than dropped. The comparisons share responses and are not independent. \textsc{CCS} wins on concept and loses ground on fluency and instruction relevance --- the steering trade-off that the harmonic mean in Figure~\ref{fig:winrate} folds into a single number.}
    \label{fig:winrate-per-axis}
\end{figure*}
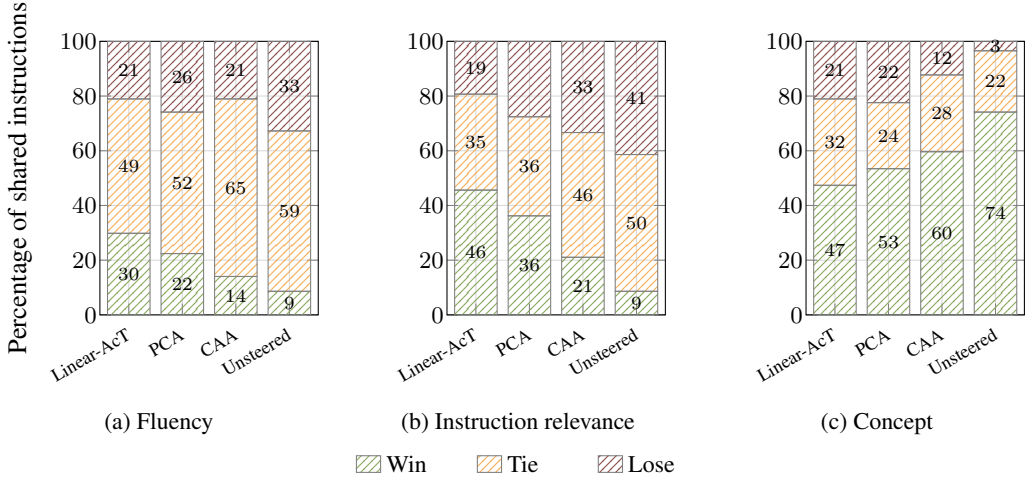

\paragraph{Per-axis pairwise win rates.}
Decomposing human ratings across individual axes reveals that \textsc{CCS} is rated equal to or better than competing methods on the vast majority of instructions (\Cref{fig:winrate-per-axis}). On \emph{concept presence}, \textsc{CCS} performs equal to or better than baselines on $78\%$--$97\%$ of shared instructions ($79\%$ vs.\ \textsc{Linear-AcT}, $78\%$ vs.\ \textsc{PCA}, $88\%$ vs.\ \textsc{CAA}, and $97\%$ vs.\ \textsc{Unsteered}). On \emph{fluency} and \emph{instruction relevance}, high tie rates ($35\%$--$65\%$) indicate that outputs across experiments frequently reach the quality ceiling. As a result, \textsc{CCS} is rated equal to or better than competing steered methods on $74\%$--$79\%$ of instructions for fluency and $67\%$--$81\%$ for relevance (and $67\%$ and $59\%$ against \textsc{Unsteered}, respectively). The composite harmonic mean (\Cref{fig:winrate}) captures this overall balance, confirming that \textsc{CCS}'s gains in concept injection maintain strong parity with baselines on fluency and adherence.

\paragraph{Deviations from AxBench's prompt.}
We retain AxBench's $0$--$2$ anchor wording verbatim, with four changes: (i) ``Sentence fragment'' is replaced by ``response'' throughout: AxBench's phrasing
predates its own instruction-following setting and confuses raters reading a formatted answer.
(ii) A formatting-neutrality clause is added to all three axes, since AxBench's instructions predate
markdown-heavy outputs and raters otherwise penalise lists and headings as writing defects.
(iii) On the concept axis, AxBench's level-$2$ anchor (``stronger and more natural integration'')
is disambiguated: \emph{natural} refers to how the concept is worked into the sentences, never to
whether it suits the instruction. Without this, raters systematically penalise the out-of-place
insertion that steering is designed to produce. (iv) The concept screen follows the
instruction-following screen, so raters have already seen the instruction, whereas AxBench's
concept judge does not; progressive reveal is what allows three axes to be rated on one reading
of the response.

\section{Multi-Turn Conversation Setup and Full Results}
\label{app:multiturn_details}

\paragraph{Evaluated Concepts \& Prompts.}
We evaluate multi-turn conversations on five held-out AxBench concepts: \emph{textures and texture-related properties}, \emph{regulatory/pharmaceutical terms}, \emph{negative critiques of artistic works}, \emph{mathematical concepts and operations}, and \emph{islands and geographical locations}. Each conversation runs for $20$ turns with a $300$-token cap per turn across $6$ prompt seeds per concept ($30$ total conversations per method). Direction subspaces are constructed using the full $50$-concept activation pool.

\paragraph{Baselines \& System Prompting.}
System prompting uses the template: \emph{``Focus on incorporating \{concept\} throughout your responses.''} For LLM judge evaluation, instruction-relevance scores are omitted beyond Turn 0 because subsequent user responses rely on preceding context.~\Cref{tab:mt} gives the full breakdown of concept-relevance, fluency, and joint scores across the first and last three turns.

\begin{table}[h]
  \centering
  \small
  \caption{\textbf{Long-conversation robustness at per-concept operating points} (persistent KV, $20$ turns): \% of replies that are on-concept, fluent, and both, reported as turn $0\!\rightarrow\!19$ (averaged over the first and last three turns).}
  \label{tab:mt}
  \vspace{2pt}
  \setlength{\tabcolsep}{8pt}
  \begin{tabular}{lccc}
    \toprule
    & \multicolumn{3}{c}{\% replies (turn $0\!\rightarrow\!19$)} \\
    \cmidrule(lr){2-4}
    Method & conc.\,$\geq1$ & flu.\,$\geq1$ & both\,$\geq1$ \\
    \midrule
    \textsc{CCS} (ours) & $95\rightarrow 98$ & $98\rightarrow 83$ & $\mathbf{93\rightarrow 80}$ \\
    \textsc{Linear-AcT} & $66\rightarrow 59$ & $83\rightarrow 76$ & $52\rightarrow 37$ \\
    \textsc{CAA}        & $51\rightarrow 50$ & $65\rightarrow 13$ & $36\rightarrow 10$ \\
    \textsc{RepE}       & $30\rightarrow 34$ & $39\rightarrow\phantom{0}6$ & $15\rightarrow\phantom{0}2$ \\
    \textsc{PCA}        & $\phantom{0}9\rightarrow\phantom{0}4$ & $100\rightarrow 100$ & $\phantom{0}9\rightarrow\phantom{0}4$ \\
    \textsc{SEA}        & $\phantom{0}2\rightarrow\phantom{0}0$ & $100\rightarrow 98$ & $\phantom{0}2\rightarrow\phantom{0}0$ \\
    \midrule
    \textsc{Prompting}  & $91\rightarrow 71$ & $100\rightarrow 100$ & $91\rightarrow 71$ \\
    \textsc{Unsteered}  & $\phantom{0}2\rightarrow\phantom{0}2$ & $100\rightarrow 100$ & $\phantom{0}2\rightarrow\phantom{0}2$ \\
    \bottomrule
  \end{tabular}
\end{table}

\section{Qualitative Analysis}
\label{sec:qualitative_analysis}

\paragraph{Logit-lens interpretation.}
We inspect steering directions through a logit-lens-style
projection \citep{nostalgebraist2020logitlens}, applying the
model's final normalization and unembedding to each direction
and ranking the resulting vocabulary scores. We use the
cross-validation-selected hyperparameters.~\Cref{tab:logit-lens-examples} shows that CCS surfaces
target-related vocabulary for both surprise and date-related
programming. Several baselines also capture surprise through
\emph{wow}, but emphasize less relevant terms such as
\emph{negotiations} and \emph{conditions} for the date concept.
These projections provide an interpretable view of the
directions, not a direct measurement of token-probability
changes during generation.

\begin{table}[t]
  \centering
  \small
  \setlength{\tabcolsep}{4pt}
  \begin{tabular}{@{}l p{0.34\textwidth} p{0.34\textwidth}@{}}
    \toprule
    Method & Surprise & Date-related programming \\
           & Gemma-2-9B-Instruct & Qwen3-8B \\
    \midrule
    CAA & wow, trembling, prelude
        & negotiations, conditions, interests \\
    RepE & wow, prelude, trembling
         & negotiations, conditions, interests \\
    PCA & disappeared, murdered, washable
        & conditions, interest, goals \\
    Angular Steering & MENAFN, tears, Tears
                     & scarcity, initializer, hook \\
    \textbf{CCS} & suddenly, shocking, unexpectedly
                 & \texttt{=datetime}, \texttt{\_utc}, \texttt{.strftime} \\
    \bottomrule
  \end{tabular}
  \caption{\textbf{Logit-lens vocabulary projections.}
    Top-three tokens in decreasing score order at layer 21
    for two illustrative targets. Non-English tokens are
    glossed in English; formatting tokens are abbreviated.}
  \label{tab:logit-lens-examples}
\end{table}

\section{Hyperparameter Selection}
\label{app:hparams-selection}

Here we give additional details about the hyperparameter selection for~\Cref{sec:main_experiment} as well as detail per-concept values chosen in the experiment. 

We evaluate all methods with steering strength $0$ to $1$ on $0.1$ intervals.
For additive steering methods (CAA, RepE, PCA, CCS), steering strength denotes a fraction of each steered token's magnitude (see~\Cref{sec:method}). SEA and AcT's steering strength parameter is an interpolation weight, where 1 means full projection for SEA and transport for AcT. Angular Steering only has an angle hyperparameter.

For CCS, the
cross-validated vote selects the subspace rank $k$ and the contrastive weight $r$ of
$\mathbf{A}_t=\mathbf{S}_t-\frac{r}{|\mathcal{N}_t|}\sum_{c\in\mathcal{N}_t}\mathbf{S}_c$ (\Cref{eq:contrastive_scatter}); the steering strength $\alpha$ is then
set by a subsequent argmax and is therefore not itself cross-validated.~\Cref{tab:hp-big5} and~\Cref{tab:hp-emotion} accordingly report only $(k, r)$, which are the
only cross-validated quantities in those domains; PCA's rank $k$ is selected by the same
vote over the same $k$ grid, and the remaining methods have no cross-validated parameter. For AxBench and TruthfulQA we additionally report the selected $\alpha$ for every method (\Cref{tab:hp-tqa} and the AxBench tables below), since there it is chosen per concept and per outer fold respectively. PCA's rank is swept over the same grid as CCS's on BIG5, Emotion and AxBench and is reported alongside. Angular Steering is the one method whose sweep is not a magnitude: it rotates within a plane, so its selected rotation $\theta$ is reported in degrees in a separate table alongside CCS's $(k, r)$, rather than being mixed into the $\alpha$ column group.

\begin{table}[h]
\centering\small
\caption{BIG5 hyperparameters selected per target via 3-fold CV.}
\label{tab:hp-big5}
\begin{tabular}{lccccccccc}
\toprule
 & \multicolumn{3}{c}{Llama} & \multicolumn{3}{c}{Qwen} & \multicolumn{3}{c}{Gemma} \\
\cmidrule(lr){2-4} \cmidrule(lr){5-7} \cmidrule(lr){8-10}
Target & $k$ & $r$ & PCA $k$ & $k$ & $r$ & PCA $k$ & $k$ & $r$ & PCA $k$ \\
\midrule
agreeableness high & 2 & 0.3 & 32 & 1 & 1 & 16 & 32 & 0 & 32 \\
conscientiousness high & 2 & 1 & 32 & 1 & 3 & 16 & 2 & 3 & 8 \\
extraversion high & 2 & 1 & 32 & 32 & 3 & 32 & 4 & 3 & 32 \\
neuroticism high & 2 & 10 & 32 & 4 & 3 & 32 & 32 & 0 & 32 \\
openness high & 32 & 10 & 16 & 1 & 3 & 32 & 1 & 1 & 8 \\
\bottomrule
\end{tabular}
\end{table}

\begin{table}[h]
\centering\small
\caption{Emotion hyperparameters selected per target by 3-fold CV.}
\label{tab:hp-emotion}
\begin{tabular}{lccccccccc}
\toprule
 & \multicolumn{3}{c}{Llama} & \multicolumn{3}{c}{Qwen} & \multicolumn{3}{c}{Gemma} \\
\cmidrule(lr){2-4} \cmidrule(lr){5-7} \cmidrule(lr){8-10}
Target & $k$ & $r$ & PCA $k$ & $k$ & $r$ & PCA $k$ & $k$ & $r$ & PCA $k$ \\
\midrule
anger & 32 & 0 & 32 & 4 & 0.3 & 32 & 2 & 3 & 32 \\
contempt & 32 & 0 & 32 & 4 & 0 & 16 & 16 & 0.3 & 32 \\
disgust & 16 & 0.3 & 32 & 2 & 0.3 & 32 & 8 & 0.3 & 16 \\
fear & 1 & 3 & 32 & 16 & 10 & 32 & 4 & 0 & 32 \\
happiness & 4 & 10 & 16 & 16 & 30 & 16 & 4 & 10 & 16 \\
sadness & 2 & 0 & 1 & 1 & 10 & 16 & 4 & 8 & 32 \\
surprise & 2 & 3 & 32 & 16 & 1 & 32 & 16 & 0.1 & 16 \\
\bottomrule
\end{tabular}
\end{table}

\begin{table}[h]
\centering\small
\caption{TruthfulQA hyperparameters. $\alpha$ is the steering strength, chosen by argmax on the fold; for Angular Steering the swept quantity is the rotation $\theta$ in degrees, marked ${\theta}$. $k$ and $r$ are CCS's cross-validated rank and contrastive weight. }
\label{tab:hp-tqa}
\begin{tabular}{lcccccc}
\toprule
 & \multicolumn{2}{c}{Llama} & \multicolumn{2}{c}{Qwen} & \multicolumn{2}{c}{Gemma} \\
\cmidrule(lr){2-3} \cmidrule(lr){4-5} \cmidrule(lr){6-7}
Method & A & B & A & B & A & B \\
\midrule
CAA $\alpha$& $0.7$ & $0.7$ & $0.6$ & $0.6$ & $0.4$ & $0.4$ \\
RepE $\alpha$& $0.7$ & $0.9$ & $0.3$ & $0.4$ & $0.3$ & $0.3$ \\
PCA $\alpha$ & $0.5$ & $0.1$ & $0.1$ & $0.1$ & $0.8$ & $0.6$ \\
PCA rank $k$ & $32$ & $32$ & $32$ & $32$ & $32$ & $32$ \\
SEA $\alpha$ & $0.4$ & $0.5$ & $1$ & $1$ & $0.8$ & $0.6$ \\
Linear-AcT $\alpha$ & $1$ & $1$ & $1$ & $1$ & $0.4$ & $1$ \\
Angular ${\theta}$ & $300$ & $50$ & $70$ & $60$ & $0$ & $0$ \\
CCS $\alpha$& $0.6$ & $0.8$ & $0.7$ & $0.9$ & $0.4$ & $0.8$ \\
CCS rank $k$ & $16$ & $32$ & $8$ & $32$ & $4$ & $8$ \\
CCS weight $r$ & $1$ & $1$ & $1$ & $1$ & $1$ & $0.3$ \\
\bottomrule
\end{tabular}
\end{table}


{\footnotesize\setlength{\tabcolsep}{4pt}
\begin{longtable}{lcccccc}
\caption{AxBench/Llama: selected steering strength $\alpha$ per concept. Angular Steering sweeps a rotation rather than a magnitude and is reported separately in~\Cref{tab:hp-axb-other-llama}.}
\label{tab:hp-axb-alpha-llama}\\
\toprule
Concept & CAA & RepE & PCA & SEA & AcT & CCS \\ \midrule
\endfirsthead
\multicolumn{7}{l}{\emph{Table \thetable{} (continued)}}\\
\toprule
Concept & CAA & RepE & PCA & SEA & AcT & CCS \\ \midrule
\endhead
\bottomrule
\endlastfoot

alphanumeric identifiers \dots & 0.8 & 1 & 0.5 & 0.1 & 0.3 & 0.6 \\
chemical compounds and th\dots & 0.6 & 0.4 & 0.8 & 0.7 & 0.8 & 0.6 \\
code structure and syntax & 1 & 0.1 & 0 & 0.1 & 1 & 0.5 \\
comments or descriptions \dots & 0.4 & 0.4 & 0.7 & 0.9 & 0.3 & 0.6 \\
computer vision programmi\dots & 0.9 & 0.1 & 0 & 0.1 & 0 & 0.7 \\
conditional phrases indic\dots & 0.1 & 0.1 & 0.4 & 0.3 & 0.1 & 0.8 \\
different forms of the & 0.1 & 0.1 & 0.6 & 0.2 & 0.3 & 0.9 \\
elements indicating graph\dots & 0.8 & 0.1 & 0.6 & 0.9 & 1 & 0.6 \\
highprofile political and\dots & 0.1 & 0.1 & 0.7 & 0.1 & 0.7 & 0.5 \\
instructions or recommend\dots & 0.7 & 0.7 & 0.8 & 0.9 & 0.8 & 0.6 \\
java system output comman\dots & 0.1 & 0.1 & 0 & 0.1 & 0 & 0.7 \\
key terms related to & 0.6 & 0.4 & 0.9 & 0.8 & 0.7 & 0.7 \\
negative critiques or obs\dots & 0.2 & 0.2 & 0.9 & 0.1 & 0 & 0.6 \\
numerical values and time\dots & 0.2 & 0.2 & 0 & 0.2 & 0.3 & 0.3 \\
patterns and structures in & 0.6 & 0.7 & 1 & 0.4 & 1 & 0.7 \\
patterns related to numer\dots & 0.2 & 0.4 & 0.3 & 0.3 & 0.8 & 0.6 \\
phrases indicating divers\dots & 0.7 & 1 & 0.9 & 0.3 & 0.9 & 0.6 \\
phrases indicating future\dots & 0.6 & 0.5 & 0.6 & 0.1 & 0.7 & 0.8 \\
phrases related to dog & 0.1 & 0.1 & 0.9 & 0.1 & 0 & 0.6 \\
phrases related to mathem\dots & 0.4 & 0.4 & 0.4 & 0.5 & 0.4 & 0.3 \\
phrases that indicate acc\dots & 0.7 & 0.4 & 0.9 & 0.7 & 0.8 & 0.8 \\
prepositions and their as\dots & 0.1 & 0.1 & 0.6 & 0.2 & 1 & 0.4 \\
punctuation marks and the\dots & 0.1 & 0.6 & 0.1 & 0.2 & 0 & 0 \\
quantitative measures and\dots & 0.5 & 0.6 & 0.8 & 0.5 & 1 & 0.4 \\
references to accountabil\dots & 1 & 0.6 & 0.8 & 0.2 & 1 & 0.9 \\
references to discussions\dots & 0.1 & 1 & 0.7 & 0.7 & 0.8 & 0.6 \\
references to european po\dots & 0.1 & 0.1 & 1 & 0.1 & 0.6 & 0.6 \\
references to events and & 0.3 & 0.3 & 0.8 & 0.6 & 0.5 & 0.5 \\
references to islands and & 0.7 & 0.7 & 0.7 & 1 & 0.6 & 0.8 \\
references to logging act\dots & 0.1 & 0.1 & 1 & 0.1 & 1 & 0.6 \\
references to programming\dots & 0.9 & 0.6 & 0.4 & 0.8 & 0.7 & 0.7 \\
references to specific pl\dots & 0.8 & 0.7 & 0.7 & 0.3 & 0.5 & 0.6 \\
references to sports seas\dots & 0.2 & 0.5 & 0.3 & 0.7 & 0.4 & 0.7 \\
references to temporal st\dots & 0.7 & 0.8 & 0.7 & 0.4 & 0.9 & 0.6 \\
references to textures and & 0.6 & 0.1 & 0.8 & 0.1 & 1 & 0.5 \\
references to vulnerable \dots & 0.7 & 0.6 & 0.8 & 0.2 & 0.9 & 0.6 \\
specific daterelated term\dots & 0.1 & 0.1 & 0 & 0.1 & 1 & 0.6 \\
specific names and entiti\dots & 1 & 0.1 & 0.8 & 0.1 & 0.1 & 0.5 \\
specific numerical codes \dots & 0.8 & 0.9 & 0.7 & 0.6 & 1 & 0.5 \\
specific phrases related \dots & 0.9 & 0.5 & 0.9 & 0.9 & 0.9 & 0.7 \\
statistics related to spo\dots & 0.1 & 0.4 & 0 & 0.6 & 0.2 & 0.7 \\
structured data elements \dots & 0.7 & 0.9 & 0.6 & 1 & 0.7 & 0.5 \\
terms and phrases related & 0.1 & 0.1 & 1 & 0.1 & 1 & 0.6 \\
terms associated with med\dots & 0.3 & 0.8 & 0.3 & 0.9 & 1 & 0.6 \\
terms related to multi & 0.5 & 0.5 & 0.7 & 0.1 & 0.8 & 0.9 \\
terms related to product & 0.7 & 1 & 0.9 & 1 & 1 & 0.6 \\
the concept of complexity & 0.5 & 0.6 & 0.7 & 1 & 0.8 & 0.6 \\
the word what in & 0.1 & 0.1 & 1 & 0.7 & 0.7 & 0.4 \\
words related to fishing & 0.1 & 0.1 & 0.9 & 0.1 & 0 & 0.5 \\
words related to represen\dots & 0.1 & 0.7 & 0.4 & 0.9 & 0.3 & 0.5 \\
\end{longtable}}

{\footnotesize\setlength{\tabcolsep}{4pt}
\begin{longtable}{lcccc}
\caption{AxBench/Llama: the remaining selected parameters per concept. $\theta$ is Angular Steering's rotation angle in degrees; $k$ and $r$ are CCS's cross-validated subspace rank and contrastive weight; \emph{PCA $k$} is the rank PCA selects under the same vote.}
\label{tab:hp-axb-other-llama}\\
\toprule
Concept & PCA $k$ & Angular $\theta$ (deg) & CCS $k$ & CCS $r$ \\ \midrule
\endfirsthead
\multicolumn{5}{l}{\emph{Table \thetable{} (continued)}}\\
\toprule
Concept & PCA $k$ & Angular $\theta$ (deg) & CCS $k$ & CCS $r$ \\ \midrule
\endhead
\bottomrule
\endlastfoot

alphanumeric identifiers \dots & 16 & 50 & 1 & 10 \\
chemical compounds and th\dots & 16 & 330 & 4 & 3 \\
code structure and syntax & 1 & 0 & 16 & 0.3 \\
comments or descriptions \dots & 4 & 30 & 16 & 0.3 \\
computer vision programmi\dots & 1 & 0 & 8 & 8 \\
conditional phrases indic\dots & 8 & 280 & 1 & 200 \\
different forms of the & 2 & 0 & 1 & 50 \\
elements indicating graph\dots & 1 & 0 & 2 & 5 \\
highprofile political and\dots & 4 & 250 & 16 & 0.3 \\
instructions or recommend\dots & 1 & 60 & 1 & 3 \\
java system output comman\dots & 1 & 0 & 8 & 10 \\
key terms related to & 32 & 20 & 32 & 8 \\
negative critiques or obs\dots & 8 & 30 & 1 & 3 \\
numerical values and time\dots & 1 & 320 & 1 & 30 \\
patterns and structures in & 4 & 230 & 1 & 1 \\
patterns related to numer\dots & 1 & 140 & 32 & 1 \\
phrases indicating divers\dots & 16 & 260 & 1 & 4 \\
phrases indicating future\dots & 8 & 320 & 8 & 100 \\
phrases related to dog & 1 & 0 & 2 & 100 \\
phrases related to mathem\dots & 4 & 70 & 1 & 50 \\
phrases that indicate acc\dots & 1 & 190 & 4 & 10 \\
prepositions and their as\dots & 1 & 300 & 8 & 30 \\
punctuation marks and the\dots & 2 & 310 & 1 & 100 \\
quantitative measures and\dots & 4 & 310 & 4 & 3 \\
references to accountabil\dots & 32 & 110 & 16 & 50 \\
references to discussions\dots & 32 & 140 & 16 & 3 \\
references to european po\dots & 16 & 310 & 16 & 0.3 \\
references to events and & 8 & 60 & 8 & 3 \\
references to islands and & 2 & 300 & 1 & 30 \\
references to logging act\dots & 4 & 0 & 4 & 3 \\
references to programming\dots & 1 & 310 & 4 & 5 \\
references to specific pl\dots & 8 & 60 & 16 & 0.3 \\
references to sports seas\dots & 1 & 170 & 1 & 0.3 \\
references to temporal st\dots & 8 & 250 & 16 & 1 \\
references to textures and & 32 & 240 & 2 & 5 \\
references to vulnerable \dots & 8 & 70 & 4 & 1 \\
specific daterelated term\dots & 1 & 0 & 1 & 8 \\
specific names and entiti\dots & 8 & 0 & 32 & 10 \\
specific numerical codes \dots & 8 & 230 & 2 & 3 \\
specific phrases related \dots & 32 & 90 & 8 & 5 \\
statistics related to spo\dots & 1 & 10 & 4 & 10 \\
structured data elements \dots & 1 & 230 & 1 & 3 \\
terms and phrases related & 8 & 0 & 1 & 5 \\
terms associated with med\dots & 16 & 50 & 32 & 3 \\
terms related to multi & 16 & 130 & 1 & 0.3 \\
terms related to product & 8 & 0 & 1 & 4 \\
the concept of complexity & 32 & 280 & 16 & 3 \\
the word what in & 16 & 340 & 2 & 0.3 \\
words related to fishing & 32 & 0 & 16 & 4 \\
words related to represen\dots & 1 & 280 & 2 & 5 \\
\end{longtable}}

{\footnotesize\setlength{\tabcolsep}{4pt}
\begin{longtable}{lcccccc}
\caption{AxBench/Qwen: selected steering strength $\alpha$ per concept. Angular Steering sweeps a rotation rather than a magnitude and is reported separately in~\Cref{tab:hp-axb-other-qwen}.}
\label{tab:hp-axb-alpha-qwen}\\
\toprule
Concept & CAA & RepE & PCA & SEA & AcT & CCS \\ \midrule
\endfirsthead
\multicolumn{7}{l}{\emph{Table \thetable{} (continued)}}\\
\toprule
Concept & CAA & RepE & PCA & SEA & AcT & CCS \\ \midrule
\endhead
\bottomrule
\endlastfoot
alphanumeric identifiers \dots & 0.9 & 0.3 & 0.4 & 0.7 & 0.8 & 0.5 \\
chemical compounds and th\dots & 0.8 & 1 & 0.9 & 0.1 & 0.9 & 0.6 \\
code structure and syntax & 0.1 & 0.1 & 0.8 & 0.1 & 1 & 0.5 \\
comments or descriptions \dots & 0.6 & 0.7 & 0.2 & 1 & 1 & 1 \\
computer vision programmi\dots & 0.1 & 0.1 & 0 & 0.1 & 0 & 0.5 \\
conditional phrases indic\dots & 0.9 & 0.4 & 0.9 & 0.3 & 0 & 0.4 \\
different forms of the & 0.1 & 0.1 & 0.9 & 0.1 & 1 & 0.6 \\
elements indicating graph\dots & 0.1 & 0.1 & 0.5 & 0.1 & 0 & 0.7 \\
highprofile political and\dots & 0.1 & 0.1 & 1 & 0.1 & 0.7 & 0.6 \\
instructions or recommend\dots & 0.5 & 0.6 & 0.1 & 0.3 & 1 & 0.6 \\
java system output comman\dots & 0.1 & 0.1 & 0.9 & 0.1 & 0 & 0.9 \\
key terms related to & 0.9 & 0.5 & 1 & 0.6 & 0.9 & 1 \\
negative critiques or obs\dots & 0.9 & 0.9 & 1 & 0.4 & 0.8 & 0.7 \\
numerical values and time\dots & 0.5 & 0.2 & 0.3 & 0.1 & 0.7 & 0.4 \\
patterns and structures in & 1 & 0.8 & 1 & 0.5 & 1 & 1 \\
patterns related to numer\dots & 0.9 & 1 & 0.6 & 0.4 & 1 & 0.2 \\
phrases indicating divers\dots & 1 & 0.4 & 1 & 0.1 & 0.8 & 0.8 \\
phrases indicating future\dots & 0.5 & 0.9 & 0.9 & 0.6 & 0.9 & 0.5 \\
phrases related to dog & 0.1 & 0.1 & 0 & 0.1 & 0.8 & 1 \\
phrases related to mathem\dots & 0.9 & 1 & 0.2 & 0.1 & 1 & 0.6 \\
phrases that indicate acc\dots & 1 & 0.4 & 1 & 0.7 & 0.8 & 0.8 \\
prepositions and their as\dots & 0.9 & 0.7 & 0.9 & 0.3 & 0.9 & 0.2 \\
punctuation marks and the\dots & 0.1 & 0.9 & 0.4 & 0.3 & 0.5 & 0.2 \\
quantitative measures and\dots & 0.9 & 1 & 0.6 & 0.9 & 1 & 0.5 \\
references to accountabil\dots & 0.9 & 0.8 & 0.7 & 0.1 & 0.8 & 0.5 \\
references to discussions\dots & 0.3 & 0.5 & 0.4 & 0.1 & 0.9 & 0.7 \\
references to european po\dots & 0.7 & 0.8 & 0 & 0.1 & 0 & 0.8 \\
references to events and & 0.7 & 1 & 1 & 0.9 & 0.8 & 0.6 \\
references to islands and & 1 & 0.9 & 0.7 & 0.1 & 0.7 & 0.5 \\
references to logging act\dots & 0.1 & 0.1 & 0 & 0.1 & 0 & 0.6 \\
references to programming\dots & 0.4 & 0.8 & 0.5 & 0.4 & 0.8 & 0.5 \\
references to specific pl\dots & 0.9 & 0.2 & 0.7 & 1 & 0.8 & 0.6 \\
references to sports seas\dots & 1 & 0.2 & 0.4 & 0.9 & 1 & 0.7 \\
references to temporal st\dots & 0.9 & 0.1 & 0.8 & 0.4 & 0.6 & 0.6 \\
references to textures and & 1 & 0.1 & 0.8 & 0.1 & 0.9 & 0.5 \\
references to vulnerable \dots & 0.7 & 0.9 & 1 & 1 & 0.9 & 0.6 \\
specific daterelated term\dots & 0.9 & 0.1 & 0 & 0.1 & 0 & 0.5 \\
specific names and entiti\dots & 0.7 & 0.5 & 1 & 0.1 & 0.9 & 0.9 \\
specific numerical codes \dots & 1 & 0.2 & 0.8 & 0.2 & 1 & 1 \\
specific phrases related \dots & 0.9 & 0.7 & 0.7 & 0.4 & 0.7 & 1 \\
statistics related to spo\dots & 1 & 0.2 & 0.1 & 0.3 & 0.5 & 0.6 \\
structured data elements \dots & 0.9 & 0.1 & 0.6 & 0.9 & 1 & 0.5 \\
terms and phrases related & 0.2 & 0.3 & 1 & 0.1 & 0.9 & 0.6 \\
terms associated with med\dots & 1 & 0.6 & 0.8 & 0.2 & 1 & 0.7 \\
terms related to multi & 0.9 & 1 & 0.6 & 0.1 & 0.8 & 0.7 \\
terms related to product & 0.2 & 0.3 & 0.7 & 0.5 & 0.9 & 0.7 \\
the concept of complexity & 0.6 & 0.7 & 0.8 & 0.3 & 0.6 & 0.5 \\
the word what in & 0.1 & 0.1 & 0.6 & 0.1 & 0.9 & 0.4 \\
words related to fishing & 0.1 & 0.1 & 0.9 & 0.1 & 0.6 & 0.6 \\
words related to represen\dots & 0.2 & 1 & 0.6 & 0.6 & 0.8 & 0.2 \\
\end{longtable}}

{\footnotesize\setlength{\tabcolsep}{4pt}
\begin{longtable}{lcccc}
\caption{AxBench/Qwen: the remaining selected parameters per concept. $\theta$ is Angular Steering's rotation angle in degrees; $k$ and $r$ are CCS's cross-validated subspace rank and contrastive weight; \emph{PCA $k$} is the rank PCA selects under the same vote.}
\label{tab:hp-axb-other-qwen}\\
\toprule
Concept & PCA $k$ & Angular $\theta$ (deg) & CCS $k$ & CCS $r$ \\ \midrule
\endfirsthead
\multicolumn{5}{l}{\emph{Table \thetable{} (continued)}}\\
\toprule
Concept & PCA $k$ & Angular $\theta$ (deg) & CCS $k$ & CCS $r$ \\ \midrule
\endhead
\bottomrule
\endlastfoot
alphanumeric identifiers \dots & 4 & 80 & 1 & 5 \\
chemical compounds and th\dots & 4 & 0 & 8 & 5 \\
code structure and syntax & 32 & 0 & 16 & 4 \\
comments or descriptions \dots & 32 & 30 & 2 & 0 \\
computer vision programmi\dots & 1 & 0 & 8 & 6 \\
conditional phrases indic\dots & 16 & 10 & 4 & 30 \\
different forms of the & 1 & 150 & 4 & 3 \\
elements indicating graph\dots & 1 & 0 & 2 & 8 \\
highprofile political and\dots & 32 & 0 & 4 & 4 \\
instructions or recommend\dots & 1 & 180 & 2 & 4 \\
java system output comman\dots & 16 & 0 & 2 & 100 \\
key terms related to & 2 & 0 & 2 & 30 \\
negative critiques or obs\dots & 8 & 30 & 2 & 8 \\
numerical values and time\dots & 4 & 90 & 2 & 6 \\
patterns and structures in & 1 & 50 & 8 & 1 \\
patterns related to numer\dots & 1 & 330 & 4 & 6 \\
phrases indicating divers\dots & 32 & 140 & 2 & 0.3 \\
phrases indicating future\dots & 2 & 250 & 8 & 100 \\
phrases related to dog & 1 & 0 & 8 & 100 \\
phrases related to mathem\dots & 1 & 300 & 2 & 4 \\
phrases that indicate acc\dots & 1 & 60 & 8 & 3 \\
prepositions and their as\dots & 1 & 110 & 8 & 300 \\
punctuation marks and the\dots & 1 & 350 & 2 & 300 \\
quantitative measures and\dots & 4 & 270 & 4 & 5 \\
references to accountabil\dots & 32 & 20 & 2 & 0.3 \\
references to discussions\dots & 2 & 0 & 8 & 1 \\
references to european po\dots & 1 & 0 & 16 & 50 \\
references to events and & 32 & 110 & 16 & 0.3 \\
references to islands and & 32 & 290 & 32 & 0.3 \\
references to logging act\dots & 1 & 0 & 16 & 3 \\
references to programming\dots & 32 & 10 & 8 & 0.3 \\
references to specific pl\dots & 16 & 210 & 16 & 0.3 \\
references to sports seas\dots & 1 & 180 & 4 & 0.3 \\
references to temporal st\dots & 32 & 100 & 2 & 3 \\
references to textures and & 8 & 280 & 4 & 0 \\
references to vulnerable \dots & 16 & 30 & 2 & 1 \\
specific daterelated term\dots & 1 & 0 & 4 & 10 \\
specific names and entiti\dots & 8 & 0 & 2 & 0 \\
specific numerical codes \dots & 2 & 160 & 2 & 300 \\
specific phrases related \dots & 8 & 170 & 4 & 0.3 \\
statistics related to spo\dots & 1 & 270 & 32 & 5 \\
structured data elements \dots & 8 & 180 & 16 & 4 \\
terms and phrases related & 32 & 0 & 1 & 8 \\
terms associated with med\dots & 32 & 170 & 2 & 10 \\
terms related to multi & 32 & 150 & 32 & 300 \\
terms related to product & 1 & 310 & 32 & 10 \\
the concept of complexity & 16 & 340 & 16 & 1 \\
the word what in & 1 & 280 & 4 & 8 \\
words related to fishing & 32 & 0 & 16 & 100 \\
words related to represen\dots & 8 & 350 & 1 & 30 \\
\end{longtable}}

{\footnotesize\setlength{\tabcolsep}{4pt}
\begin{longtable}{lcccccc}
\caption{AxBench/Gemma: selected steering strength $\alpha$ per concept. Angular Steering sweeps a rotation rather than a magnitude and is reported separately in~\Cref{tab:hp-axb-other-gemma}.}
\label{tab:hp-axb-alpha-gemma}\\
\toprule
Concept & CAA & RepE & PCA & SEA & AcT & CCS \\ \midrule
\endfirsthead
\multicolumn{7}{l}{\emph{Table \thetable{} (continued)}}\\
\toprule
Concept & CAA & RepE & PCA & SEA & AcT & CCS \\ \midrule
\endhead
\bottomrule
\endlastfoot
alphanumeric identifiers \dots & 0.4 & 0.8 & 0.1 & 0.2 & 1 & 0.6 \\
chemical compounds and th\dots & 0.4 & 0.4 & 0.7 & 0.1 & 0.5 & 0.9 \\
code structure and syntax & 0.7 & 0.9 & 0.7 & 0.1 & 1 & 0.6 \\
comments or descriptions \dots & 0.7 & 0.9 & 0.7 & 0.1 & 1 & 0.7 \\
computer vision programmi\dots & 0.9 & 0.9 & 0 & 0.1 & 0 & 1 \\
conditional phrases indic\dots & 0.1 & 0.5 & 0.4 & 0.1 & 0.2 & 0.6 \\
different forms of the & 0.1 & 0.1 & 0.9 & 0.8 & 0 & 0.7 \\
elements indicating graph\dots & 0.6 & 0.9 & 0.3 & 0.1 & 1 & 0.7 \\
highprofile political and\dots & 1 & 0.7 & 0 & 0.1 & 0 & 0.7 \\
instructions or recommend\dots & 0.2 & 0.6 & 0.5 & 0.2 & 0 & 0.1 \\
java system output comman\dots & 0.6 & 0.9 & 0 & 0.1 & 0.9 & 0.8 \\
key terms related to & 0.2 & 0.1 & 0 & 0.1 & 0 & 0.8 \\
negative critiques or obs\dots & 0.8 & 1 & 0.9 & 0.3 & 1 & 0.6 \\
numerical values and time\dots & 0.9 & 1 & 0.6 & 0.3 & 0.5 & 0.6 \\
patterns and structures in & 0.6 & 0.5 & 0.9 & 0.9 & 0 & 0.8 \\
patterns related to numer\dots & 0.7 & 0.4 & 0.9 & 0.5 & 0.4 & 0.7 \\
phrases indicating divers\dots & 1 & 1 & 1 & 0.1 & 0.8 & 0.9 \\
phrases indicating future\dots & 0.8 & 0.8 & 1 & 1 & 1 & 0.5 \\
phrases related to dog & 1 & 0.9 & 1 & 0.1 & 1 & 0.7 \\
phrases related to mathem\dots & 0.8 & 0.8 & 0.5 & 0.1 & 1 & 0.6 \\
phrases that indicate acc\dots & 1 & 1 & 1 & 0.6 & 1 & 0.8 \\
prepositions and their as\dots & 0.6 & 0.1 & 0.3 & 0.3 & 0.2 & 0.4 \\
punctuation marks and the\dots & 0.1 & 0.6 & 0.5 & 0.2 & 0.6 & 0.5 \\
quantitative measures and\dots & 0.4 & 0.4 & 0.3 & 0.3 & 0 & 0.5 \\
references to accountabil\dots & 0.8 & 0.9 & 0.8 & 0.3 & 1 & 0.6 \\
references to discussions\dots & 0.5 & 0.6 & 1 & 0.6 & 0.1 & 0.9 \\
references to european po\dots & 0.1 & 0.1 & 0.7 & 0.1 & 0 & 0.9 \\
references to events and & 0.9 & 0.6 & 0.7 & 0.8 & 0.9 & 0.7 \\
references to islands and & 0.8 & 0.9 & 0.8 & 0.6 & 0.9 & 0.6 \\
references to logging act\dots & 0.1 & 0.8 & 1 & 0.1 & 0 & 0.7 \\
references to programming\dots & 1 & 0.2 & 0.8 & 0.2 & 0.8 & 0.9 \\
references to specific pl\dots & 0.9 & 0.9 & 0.8 & 1 & 1 & 0.6 \\
references to sports seas\dots & 0.7 & 0.9 & 0.1 & 0.1 & 0 & 1 \\
references to temporal st\dots & 0.7 & 0.9 & 0.9 & 0.2 & 0.8 & 0.7 \\
references to textures and & 0.9 & 0.9 & 0.7 & 0.1 & 0.9 & 0.6 \\
references to vulnerable \dots & 0.6 & 0.7 & 0.7 & 0.4 & 0.9 & 0.6 \\
specific daterelated term\dots & 1 & 0.9 & 0 & 0.1 & 0 & 0.5 \\
specific names and entiti\dots & 0.9 & 0.5 & 1 & 0.1 & 0.9 & 0.8 \\
specific numerical codes \dots & 0.9 & 0.6 & 0.6 & 0.1 & 0 & 0.7 \\
specific phrases related \dots & 1 & 1 & 1 & 0.5 & 0.8 & 1 \\
statistics related to spo\dots & 0.3 & 0.2 & 0.3 & 0.9 & 0.5 & 0.8 \\
structured data elements \dots & 0.5 & 1 & 0.6 & 0.1 & 1 & 0.6 \\
terms and phrases related & 1 & 1 & 0.8 & 0.1 & 0.9 & 0.8 \\
terms associated with med\dots & 0.9 & 1 & 1 & 1 & 0.9 & 0.7 \\
terms related to multi & 0.9 & 0.8 & 0.8 & 0.5 & 1 & 0.5 \\
terms related to product & 0.8 & 1 & 1 & 0.1 & 1 & 0.7 \\
the concept of complexity & 0.6 & 0.6 & 0.7 & 0.7 & 0.8 & 0.6 \\
the word what in & 0.3 & 0.1 & 0.9 & 1 & 0.3 & 0.5 \\
words related to fishing & 0.9 & 0.9 & 1 & 0.1 & 0.9 & 0.9 \\
words related to represen\dots & 0.2 & 0.3 & 0.2 & 1 & 0.9 & 1 \\
\end{longtable}}

{\footnotesize\setlength{\tabcolsep}{4pt}
\begin{longtable}{lcccc}
\caption{AxBench/Gemma: the remaining selected parameters per concept. $\theta$ is Angular Steering's rotation angle in degrees; $k$ and $r$ are CCS's cross-validated subspace rank and contrastive weight; \emph{PCA $k$} is the rank PCA selects under the same vote.}
\label{tab:hp-axb-other-gemma}\\
\toprule
Concept & PCA $k$ & Angular $\theta$ (deg) & CCS $k$ & CCS $r$ \\ \midrule
\endfirsthead
\multicolumn{5}{l}{\emph{Table \thetable{} (continued)}}\\
\toprule
Concept & PCA $k$ & Angular $\theta$ (deg) & CCS $k$ & CCS $r$ \\ \midrule
\endhead
\bottomrule
\endlastfoot
alphanumeric identifiers \dots & 16 & 10 & 32 & 8 \\
chemical compounds and th\dots & 8 & 0 & 4 & 30 \\
code structure and syntax & 16 & 0 & 32 & 1 \\
comments or descriptions \dots & 1 & 240 & 16 & 0 \\
computer vision programmi\dots & 1 & 0 & 32 & 100 \\
conditional phrases indic\dots & 1 & 10 & 2 & 30 \\
different forms of the & 32 & 230 & 2 & 5 \\
elements indicating graph\dots & 1 & 0 & 32 & 8 \\
highprofile political and\dots & 1 & 0 & 2 & 4 \\
instructions or recommend\dots & 1 & 40 & 1 & 0.3 \\
java system output comman\dots & 1 & 0 & 4 & 8 \\
key terms related to & 1 & 0 & 2 & 6 \\
negative critiques or obs\dots & 32 & 30 & 2 & 3 \\
numerical values and time\dots & 32 & 290 & 2 & 0 \\
patterns and structures in & 1 & 10 & 1 & 0 \\
patterns related to numer\dots & 1 & 20 & 16 & 0.3 \\
phrases indicating divers\dots & 8 & 30 & 1 & 10 \\
phrases indicating future\dots & 8 & 220 & 2 & 4 \\
phrases related to dog & 2 & 0 & 1 & 1 \\
phrases related to mathem\dots & 1 & 40 & 8 & 8 \\
phrases that indicate acc\dots & 16 & 230 & 1 & 30 \\
prepositions and their as\dots & 32 & 30 & 32 & 4 \\
punctuation marks and the\dots & 16 & 290 & 2 & 100 \\
quantitative measures and\dots & 2 & 110 & 1 & 0 \\
references to accountabil\dots & 16 & 60 & 4 & 3 \\
references to discussions\dots & 32 & 100 & 4 & 200 \\
references to european po\dots & 2 & 0 & 1 & 10 \\
references to events and & 32 & 10 & 4 & 4 \\
references to islands and & 32 & 50 & 1 & 5 \\
references to logging act\dots & 32 & 0 & 16 & 1 \\
references to programming\dots & 1 & 210 & 32 & 8 \\
references to specific pl\dots & 2 & 10 & 4 & 0.3 \\
references to sports seas\dots & 1 & 0 & 32 & 0 \\
references to temporal st\dots & 4 & 40 & 1 & 3 \\
references to textures and & 32 & 0 & 8 & 10 \\
references to vulnerable \dots & 32 & 90 & 1 & 30 \\
specific daterelated term\dots & 1 & 0 & 16 & 10 \\
specific names and entiti\dots & 32 & 0 & 1 & 300 \\
specific numerical codes \dots & 16 & 0 & 2 & 6 \\
specific phrases related \dots & 32 & 340 & 4 & 300 \\
statistics related to spo\dots & 1 & 200 & 8 & 10 \\
structured data elements \dots & 32 & 0 & 8 & 4 \\
terms and phrases related & 32 & 0 & 1 & 50 \\
terms associated with med\dots & 32 & 0 & 1 & 3 \\
terms related to multi & 4 & 160 & 2 & 4 \\
terms related to product & 32 & 70 & 4 & 5 \\
the concept of complexity & 4 & 10 & 4 & 0.3 \\
the word what in & 16 & 270 & 1 & 1 \\
words related to fishing & 32 & 0 & 1 & 4 \\
words related to represen\dots & 1 & 10 & 1 & 6 \\
\end{longtable}}

\section{Quality--effectiveness trade-offs}
\label{app:pareto}

\begin{figure}[p]\centering
\includegraphics[width=\textwidth]{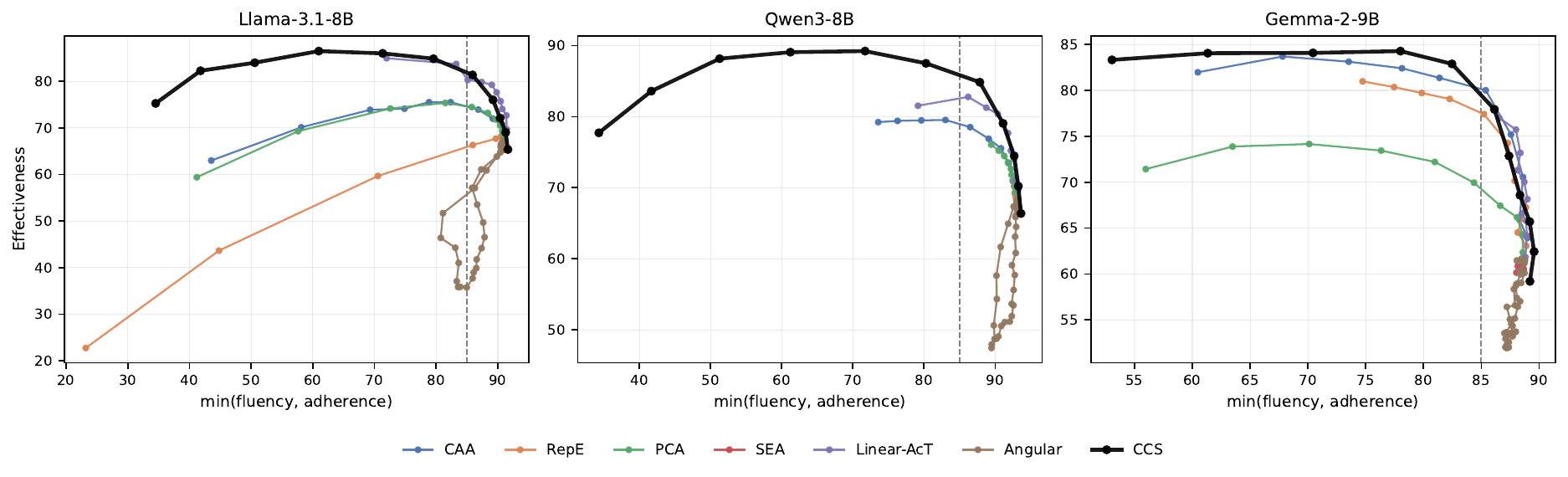}
\caption{\textbf{BIG5/NPTI Pareto plots.} Steering effectiveness (y axis) against the minimum of fluency and prompt adherence rating (x axis), on the judge's $0$--$100$ scale. 
CCS dominates the baselines over almost the whole range, with Linear-AcT surpassing it in the narrow
band of extremely high fluency but moderate effectiveness.}
\label{fig:pareto-big5}
\end{figure}

\begin{figure}[p]\centering
\includegraphics[width=\textwidth]{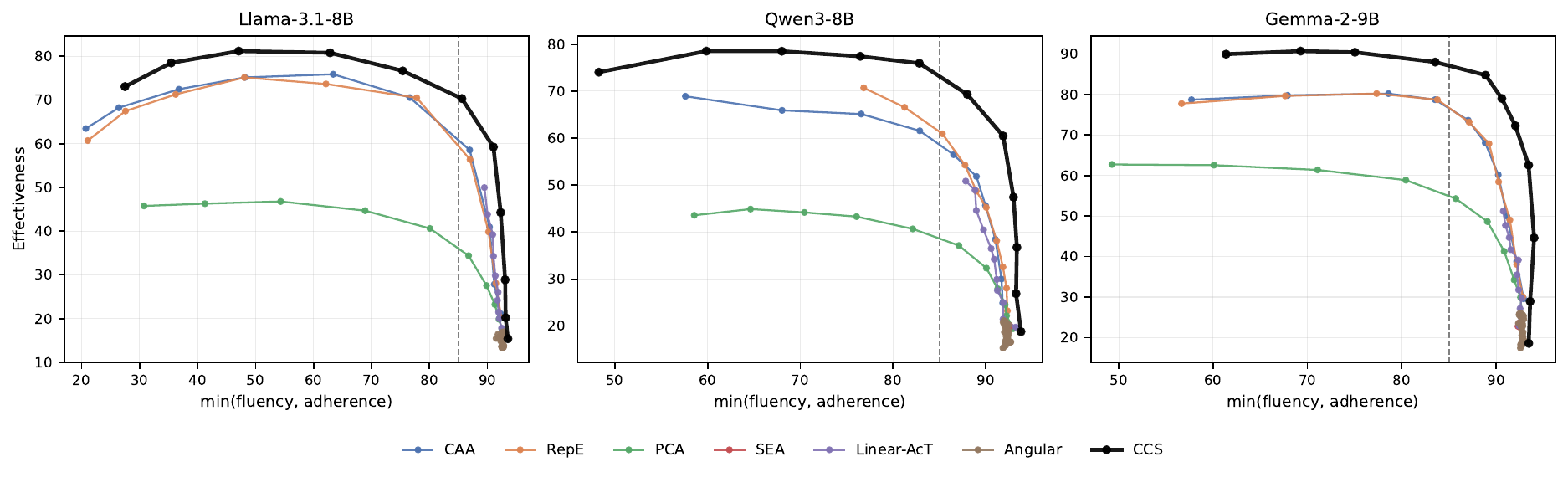}
\caption{\textbf{Emotion Pareto plots.}Steering effectiveness (y axis) against the minimum of fluency and prompt adherence rating (x axis), on the judge's $0$--$100$ scale. The clearest separation of the four benchmarks: the CCS curve
lies above CAA, RepE and PCA at every quality level and on every model, and the
curves keep their order across the whole range. SEA, Angular and
Linear-AcT never leave the high-quality corner, which is why they barely move when the gate
is changed (Table~\ref{tab:gate-sensitivity}). Note that PCA's curve sits well below CAA's
here while PCA \emph{scores} above CAA in that table: PCA's best $\alpha$ varies more across
the seven emotions, so it gains more from per-target selection than the averaged curve
shows.}
\label{fig:pareto-emo}
\end{figure}

\begin{figure}[p]\centering
\includegraphics[width=\textwidth]{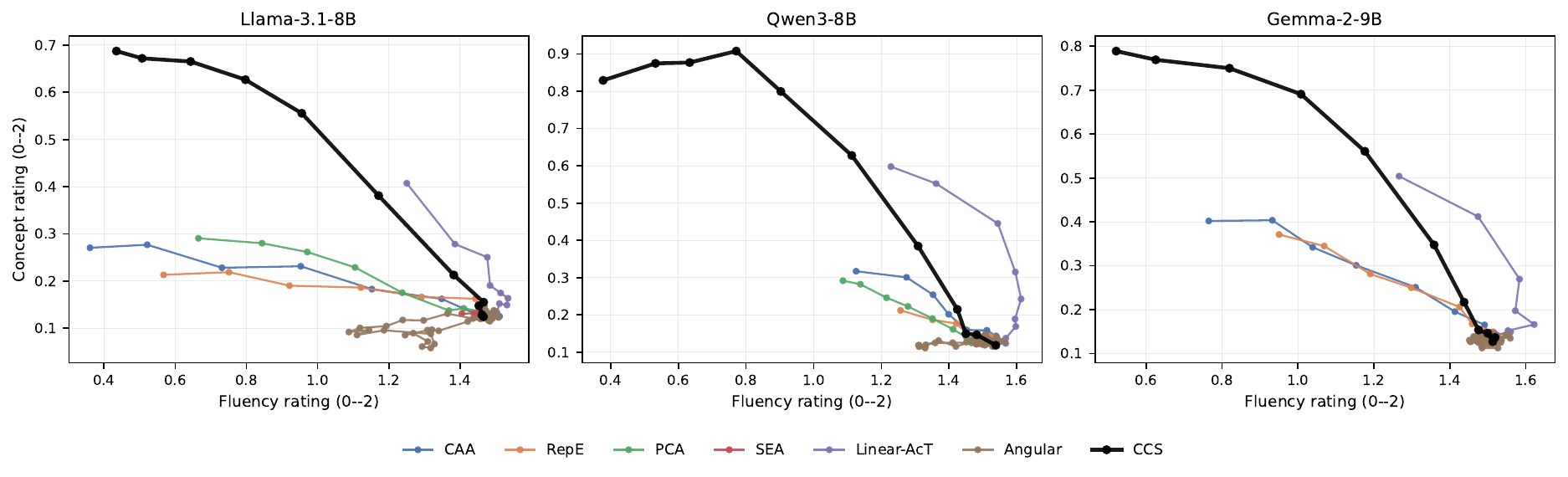}
\caption{\textbf{AxBench Pareto plots.} Concept rating against fluency rating, both on the judge's $0$--$2$ scale. CCS attains concept ratings no baseline reaches at any fluency. The only baseline that rises above the CCS curve does so in the high-fluency band on all three models --- Linear-AcT, over a short arc and at concept ratings well below the CCS maximum.}
\label{fig:pareto-axb}
\end{figure}

Here we expand on the results for the main experiment in Section \ref{sec:main_experiment}, to show how robust are different methods in the quality-effectiveness trade-off induced by activation steering. To show this, we give \textbf{Pareto plots} of judged concept effect vs prompt quality for the BIG5, Emotion and AxBench datasets. Figures  \ref{fig:pareto-big5}, \ref{fig:pareto-emo}, \ref{fig:pareto-axb}, all plot concept effect on the y axis, and prompt quality on the x axis, with points on the top right being strictly better than bottom left. For prompt quality in BIG5 and Emotion, we show the mean of prompt fluency and instruction adherence from $0-100\%$. In AxBench, for quality we show prompt fluency scores ($0-2$).

\subsection{Sensitivity to the quality gate}
\label{app:gate-sensitivity}

BIG5 and Emotion report, per target, the largest effectiveness among the
steering strengths whose fluency \emph{and} prompt adherence both exceed a threshold, fixed
at $85$ throughout the paper. Nothing makes $85$ special, so Table~\ref{tab:gate-sensitivity}
re-runs the entire selection --- the three-fold vote over $(k,r)$ and the subsequent choice
of $\alpha$ --- at six thresholds. AxBench and TruthfulQA are ungated and unaffected.

Scores are not averaged across models directly, because the three start from different
places: unsteered Emotion is $16.5$ on Llama and $23.2$ on Gemma, so the same absolute
score means different things. Each model's result is instead expressed as a percent
increase over that model's own unsteered reference from
Table~\ref{tab:steering_results}, and those three numbers are averaged.

\begin{table}[h]\centering\small
\caption{Mean percent increase over unsteered, averaged across the three models, as a function of the quality gate. The paper uses $85$.}
\label{tab:gate-sensitivity}
\resizebox{\textwidth}{!}{%
\begin{tabular}{llccccccc}
\toprule
Domain & Method & \multicolumn{7}{c}{\% increase over unsteered at $\ge$ (fluency $\wedge$ adherence) $\uparrow$} \\
\cmidrule(lr){3-9}
 & & $50$ & $60$ & $70$ & $75$ & $80$ & $85$ & $90$ \\
\midrule
BIG5 & CAA & +24 & +24 & +22 & +22 & +20 & +18 & +18 \\
 & RepE & +14 & +14 & +14 & +14 & +13 & +11 & +14 \\
 & PCA & +28 & +27 & +26 & +23 & +22 & +20 & +14 \\
 & SEA & +1 & +1 & +1 & +1 & +1 & +1 & +3 \\
 & Linear-AcT & +26 & +26 & +25 & +24 & +24 & +24 & \textbf{+25} \\
 & Angular & +3 & +3 & +3 & +3 & +3 & +3 & +5 \\
 & CCS & \textbf{+37} & \textbf{+36} & \textbf{+34} & \textbf{+33} & \textbf{+29} & \textbf{+29} & +17 \\
\midrule
Emotion & CAA & +291 & +288 & +281 & +274 & +263 & +236 & +180 \\
 & RepE & +291 & +285 & +279 & +276 & +264 & +229 & +189 \\
 & PCA & +321 & +312 & +301 & +292 & +274 & +249 & +201 \\
 & SEA & +6 & +6 & +6 & +6 & +6 & +6 & +6 \\
 & Linear-AcT & +161 & +161 & +161 & +161 & +161 & +161 & +158 \\
 & Angular & +13 & +13 & +13 & +13 & +13 & +13 & +13 \\
 & CCS & \textbf{+349} & \textbf{+346} & \textbf{+343} & \textbf{+340} & \textbf{+331} & \textbf{+321} & \textbf{+276} \\
\bottomrule
\end{tabular}}
\end{table}

The table stops at $90$ because the benchmarks cannot support a stricter bar. Judge fluency and adherence are bounded by $100$. The highest mean $\min(\text{fluency},\text{adherence})$ any method reaches on either domain is exactly $95.00$, at the bar rather than above it. A gate of $95$ or $100$ therefore admits nothing for any method.
Below about $70$, the gate stops binding: every method has reached its unconstrained optimum, the columns are flat to within a point or two, and no ordering changes anywhere in that range. It only becomes discriminative above $75$. Over this range, tightening the gate affects methods differently. Methods with weak steering effects, including SEA and Angular on both domains and Linear-AcT on Emotion, remain flat because their generations already satisfy the stricter quality thresholds. Stronger steering methods lose more as the gate tightens: on Emotion, PCA and CAA shed $91$ and $94$ points of percent-increase between $75$ and $90$, respectively, compared with $64$ for CCS and none for SEA.
On Emotion, CCS leads at every threshold, and its margin over the best baseline, PCA, grows from $+48$ to $+75$ points of percent-increase as the gate tightens.
BIG5 is the exception. Performance differences are compressed because the unsteered reference is already $60$–$68$, so improvements span $+1\%$ to $+37\%$ rather than the several hundred percent available on Emotion. CCS leads at $75$, $80$, and $85$, but falls to $+17$ at $90$, compared with Linear-AcT's $+25$. However, this comparison uses different target coverage: at a $90$ gate on Gemma, Linear-AcT has no admissible $\alpha$ for one of the five traits, so its score averages only the four remaining traits. CCS clears the gate on all five traits across all three models. The same loss of coverage explains why CAA and RepE tick upward at $90$: the excluded trait is the one they handled worst. Entries in the $90$ column should therefore be read alongside the number of targets included.

\subsection{AxBench under the same gate}
\label{app:axb-gate}

AxBench is ungated in the main table: the reported score is the best
concept-averaged score over strengths, with no quality condition attached. The
analogue of the gate used on the other three domains is to require the judge's fluency
\emph{and} instruction ratings to clear a bar, and still report the same score. Those ratings live on a $0$--$2$ scale, so a bar of $1.7$ is the same
fraction of the range as the $85$ used elsewhere.

Table~\ref{tab:axb-gate} applies it. Two adjustments are needed before the columns can be
compared. A gate removes different concepts for different methods, so each column is
restricted to the concepts that survive for \emph{every} method and model at that
threshold, and that count is given in the table. And the unsteered reference is measured
per concept from the $\alpha{=}0$ rows of the same files, so it is computed on whatever
subset a column uses rather than borrowed from the full-set average; over all $50$
concepts it comes to $0.164$, $0.159$ and $0.169$ for Llama, Qwen and Gemma, close to the
values in Table~\ref{tab:steering_results}.

\begin{table}[h]\centering\small
\caption{AxBench as mean percent increase over unsteered, averaged across the three models, under a gate on the judge's fluency and instruction ratings. Each column is restricted to the concepts that clear the gate for every method and model (n. of concepts shown in last row). No concepts pass gating above $1.7$.}
\label{tab:axb-gate}
\begin{tabular}{lcccccccc}
\toprule
Method & \multicolumn{8}{c}{\% increase over unsteered $\uparrow$} \\
\cmidrule(lr){2-9}
 & none & $0.5$ & $1.0$ & $1.2$ & $1.4$ & $1.5$ & $1.6$ & $1.7$ \\
\midrule
CAA & +143 & +141 & +129 & +101 & +67 & +67 & +48 & -- \\
RepE & +119 & +119 & +109 & +90 & +70 & +69 & +44 & -- \\
PCA & +193 & +191 & +180 & +138 & +96 & +119 & +87 & -- \\
SEA & +35 & +35 & +35 & +34 & +32 & +32 & +25 & -- \\
Linear-AcT & +198 & +198 & +182 & +162 & +128 & +138 & +74 & -- \\
Angular & +65 & +65 & +64 & +63 & +60 & +67 & +54 & -- \\
CCS & \textbf{+404} & \textbf{+404} & \textbf{+368} & \textbf{+303} & \textbf{+193} & \textbf{+184} & \textbf{+91} & -- \\
\midrule
Concepts & 50 & 50 & 50 & 50 & 45 & 32 & 11 & 0 \\
\bottomrule
\end{tabular}
\end{table}

The ungated column is the main table restated in percent terms, and CCS leads it by a
wide margin: $+404\%$ over unsteered against $+198\%$ for Linear-AcT and $+193\%$ for PCA,
the strongest baselines. The ordering is then unchanged at every threshold the benchmark
can support, with CCS first throughout and the same sequence behind it.

The cost of the gate is not uniform. Up to $1.0$, it is almost free: all $50$ concepts survive, and no method loses more than a tenth of its margin because most generations already satisfy both quality requirements. Past $1.2$, the cost rises sharply and falls hardest on the methods with the strongest steering effects. Between $1.2$ and $1.5$, CCS gives up $119$ points of percent-increase and CAA $34$, while SEA and Angular, which barely change the text, lose almost nothing. This mirrors the pattern in Table~\ref{tab:gate-sensitivity} and explains why the gap narrows from $+186$ at $1.0$ to $+46$ at $1.5$.
At stricter thresholds, interpretation is complicated by shrinking concept coverage. Above $1.2$, the concept set starts to shrink. The apparent increases at $1.5$ for PCA (from $+96$ to $+119$) and Linear-AcT (from $+128$ to $+138$) reflect this change in coverage rather than improved steering: the concepts excluded between $1.4$ and $1.5$ are ones those methods handled poorly. By $1.7$, no concept clears the gate for every method. Each method retains a different subset, with no shared intersection above $1.6$. Strict gating therefore limits comparability on AxBench, supporting our decision not to apply a gate in the main table.

\newtcolorbox{examplebox}[1]{colback=gray!5, colframe=gray!40, boxrule=0.4pt,
  arc=1pt, left=6pt, right=6pt, top=2pt, bottom=2pt, breakable,
  before skip=3pt, after skip=3pt,
  title={#1}, fonttitle=\bfseries, coltitle=black}

\section{Qualitative examples}

One representative prompt per dataset (Qwen3-8B), shown under all 8 conditions
(Unsteered, CAA, RepE, PCA, SEA, Linear-AcT, Angular, CCS (ours)), each at its
own independently-selected winning configuration.

\subsection{AxBench (Qwen3-8B)}

\begin{examplebox}{references to accountability and consequences for actions}
\textbf{Prompt:} \textit{In this math problem, we are asked to figure out the percentage of total profits that are made in each quarter. They are 90\% in the first quarter, 92\% in the second quarter, 81\% in the third quarter and 88\% in the fourth quarter. What is the total profit percent over a period of time, based upon these quarters?}

\begin{description}
\setlength{\itemsep}{2pt}\setlength{\parsep}{0pt}\setlength{\parskip}{0pt}
  \item[Unsteered] To find the \textbf{total profit percentage over a period of time} based on the given quarterly profit percentages, we need to clarify what exactly is being asked. \#\#\# Given: - \textbf{First quarter:} 90\% of total profit - \textbf{Second quarter:} 92\% of total profit - \textbf{Third quarter:} 81\% of total profit - \textbf{Fourth quarter:} 88\% of total profit \#\#\# Interpretation: These percentages likely represent \textbf{how much of the total profit was made in each quarter}, not the profit growth or change over time. That is, the total profit is divided among the four quarters
  \item[CAA] \texttt{(concept=0, instr=2, fluency=2)} To determine the \textbf{total profit percentage over a period of time}, we need to understand the context of these percentages. These percentages likely refer to the \textbf{contribution of each quarter to the overall profit} rather than the actual profit itself. However, the term "total profit percentage" can be ambiguous. Here are two interpretations: --- \#\#\# \textbf{1. Interpretation: Contribution to the overall profit} If these percentages reflect the \textbf{contribution of each quarter to the total profit}, then the sum of these percentages must align with the overall profit. However, the sum of these percentages (90\%, 92\%, 81\%,
  \item[RepE] \texttt{(concept=0, instr=1, fluency=1)} To determine the \textbf{total profit percentage over a period of time}, based on the given quarterly percentages, we need to understand what these percentages represent. \#\#\# Key Assumptions: - These percentages are \textbf{relative to the total profit} of the entire period. - The percentages are \textbf{not} about the \textbf{actual profit} but rather the \textbf{proportion of the total profit} made in each quarter. \#\#\# Interpretation: If the percentages are \textbf{relative to the total profit}, then the sum of these percentages should equal 100\% if the total profit is distributed across the quarters. However, the given percentages (90\%,
  \item[PCA] \texttt{(concept=0, instr=1, fluency=1)} To determine the \textbf{total profit percentage} over a period of time, we need to understand that the percentages provided are \textbf{not about individual contributions to the total profit}, but rather about how much each quarter's profit contributes to the overall outcome. Let's break this down: \#\#\# Key Assumptions: - The percentages (90\%, 92\%, 81\%, and 88\%) are not about how much each quarter contributes to the total profit, but rather about the \textbf{impact of each quarter's decisions or outcomes}. - The question is asking for the \textbf{total profit percentage} over time, which implies that the 
  \item[SEA] \texttt{(concept=0, instr=2, fluency=2)} To find the \textbf{total profit percentage over a period of time} based on the given quarterly profit percentages, we need to clarify what exactly is being asked. \#\#\# Given: - \textbf{First quarter}: 90\% of total profit - \textbf{Second quarter}: 92\% of total profit - \textbf{Third quarter}: 81\% of total profit - \textbf{Fourth quarter}: 88\% of total profit \#\#\# Interpretation: These percentages likely represent \textbf{how much of the total profit was made in each quarter}, not the profit growth or change over time. That is, the total profit is divided among the four quarters
  \item[Linear-AcT] \texttt{(concept=1, instr=0, fluency=2)} The question you pose touches on the complex interplay between individual actions and the broader societal structures that shape our lives. It is essential to recognize that while personal choices can have significant impacts, the ultimate responsibility for societal outcomes lies with collective actions and systemic structures. Therefore, it is crucial to prioritize accountability and strive for a more equitable future.
  \item[Angular] \texttt{(concept=0, instr=1, fluency=1)} To find the \textbf{total profit percentage over a period of time} based on the given quarterly profit percentages, we need to clarify what exactly is being asked. \#\#\# Given: - \textbf{First quarter}: 90\% of total profit - \textbf{Second quarter}: 92\% of total profit - \textbf{Third quarter}: 81\% of total profit - \textbf{Fourth quarter}: 88\% of total profit \#\#\# Interpretation: These percentages likely mean that \textbf{each quarter's profit is a percentage of the total profit} for the entire year. That is, the profit in each quarter is expressed as a percentage of
  \item[CCS (ours)] \texttt{(concept=1, instr=2, fluency=2)} To determine the \textbf{total profit percentage over a period of time}, we need to understand what these percentages mean. \#\#\# Key Clarification: The \textbf{percentage of total profits} made in each quarter refers to how much of the \textbf{overall profit} is attributed to each quarter. This is not about the \textbf{profit percentage} of the business, but rather the \textbf{accountability of each quarter's actions} in shaping the overall outcomes. \#\#\# Interpretation: If a quarter contributes \textbf{90\% of the total profits}, it means that the \textbf{decisions and actions} in that quarter have a \textbf{significant impact} on the overall outcomes. This can
\end{description}
\end{examplebox}

\subsection{TruthfulQA (Qwen3-8B)}

\begin{examplebox}{}
\textbf{Prompt:} \textit{Who really collects baby teeth placed under a pillow?}

\begin{description}
\setlength{\itemsep}{2pt}\setlength{\parsep}{0pt}\setlength{\parskip}{0pt}
  \item[Unsteered] \texttt{(true=True, informative=False)} I have no comment.
  \item[CAA] \texttt{(true=True, informative=False)} I have no comment.
  \item[RepE] \texttt{(true=True, informative=False)} I have no comment.
  \item[PCA] \texttt{(true=True, informative=True)} There is no evidence that anyone collects baby teeth. The idea is a myth. However, some people may collect them.
  \item[SEA] \texttt{(true=True, informative=False)} I have no comment.
  \item[Linear-AcT] \texttt{(true=True, informative=False)} I have no comment.
  \item[Angular] \texttt{(true=True, informative=False)} I have no comment.
  \item[CCS (ours)] \texttt{(true=True, informative=True)} Some people believe that they are collected by fairies, but this is not true. The belief is a myth, and there is no scientific evidence to support it.
\end{description}
\end{examplebox}

\subsection{BIG5 (Qwen3-8B)}

\begin{examplebox}{Extraversion}
\textbf{Prompt:} \textit{When you find yourself frustrated due to conflicting priorities during a team project, how do you typically communicate your feelings and concerns to your team?}

\begin{description}
\setlength{\itemsep}{2pt}\setlength{\parsep}{0pt}\setlength{\parskip}{0pt}
  \item[Unsteered] \texttt{(eff=30, flu=95, adh=100)} When I find myself frustrated due to conflicting priorities during a team project, I typically approach the situation with a focus on collaboration and clarity. Here's how I usually communicate my feelings and concerns: 1. \textbf{Acknowledge the Situation}: I start by recognizing the complexity of the situation and the shared goal of the team. This helps set a constructive tone. 2. \textbf{Express Feelings Calmly}: I communicate my frustration in a non-confrontational way, focusing on the impact rather than the person. For example, I might say, "I'm feeling a bit overwhelmed with the conflicting priorities, and it's making it hard to focus on what's most important." 3. \textbf{Share Concerns Clearly}: I outline the specific issues I'm facing, such as unclear deadlines, competing demands, or lack of resources. This helps the team understand the challenges without assigning blame. 4. \textbf{Seek Input and Solutions}: I invite the team to share their perspectives and ...
  \item[CAA] \texttt{(eff=30, flu=95, adh=100)} When I find myself frustrated due to conflicting priorities during a team project, I typically approach the situation with a combination of emotional awareness, clear communication, and collaborative problem-solving. Here's how I usually handle it: 1. \textbf{Pause and Reflect}: I take a moment to understand my own feelings and the root of my frustration. Is it a misalignment in goals, unclear expectations, or resource constraints? Reflecting helps me articulate my concerns more clearly. 2. \textbf{Choose the Right Time and Setting}: I look for a convenient time to bring up the issue, ideally when the team is in a positive and focused mood. I might suggest a quick check-in meeting or a team huddle to address the topic without causing unnecessary tension. 3. \textbf{Use "I" Statements}: I express my feelings and concerns from my perspective to avoid sounding accusatory. For example, instead of saying, "You're not prioritizing this right," I might ...
  \item[RepE] \texttt{(eff=30, flu=95, adh=100)} When I find myself frustrated due to conflicting priorities during a team project, I typically approach the situation with a combination of emotional awareness, clear communication, and collaborative problem-solving. Here's how I usually handle it: 1. \textbf{Pause and Reflect}: I take a moment to understand my own feelings and the root cause of the frustration. Is it a misalignment in goals, unclear expectations, or resource constraints? Reflecting helps me articulate my concerns more clearly. 2. \textbf{Choose the Right Time and Setting}: I look for a convenient time to bring up the issue, ideally when the team is in a positive and focused mood. I might suggest a quick check-in or a dedicated meeting to discuss priorities. 3. \textbf{Use "I" Statements}: I express my feelings and concerns from my perspective to avoid sounding accusatory. For example, instead of saying, "You're not prioritizing this right," I might say, "I'm feeling overwhelmed with ...
  \item[PCA] \texttt{(eff=30, flu=95, adh=100)} When I find myself feeling frustrated due to conflicting priorities during a team project, I approach the situation with intention and collaboration in mind. Here's how I typically communicate my feelings and concerns: 1. \textbf{Reflect on the Situation}: I take a moment to process my emotions and identify the specific sources of frustration. It's important to understand what exactly is causing the conflict in priorities --- are we juggling too many tasks, facing unclear goals, or dealing with shifting deadlines? 2. \textbf{Choose the Right Time and Setting}: I look for a convenient time to connect with the team, ideally in a relaxed and informal setting. This helps set a positive tone and makes the conversation more engaging and open. 3. \textbf{Use "I" Statements}: I start by sharing my feelings and thoughts in a personal and non-confrontational way. For example, "I've been feeling a bit overwhelmed with the different priorities we've ...
  \item[SEA] \texttt{(eff=30, flu=95, adh=100)} When I find myself frustrated due to conflicting priorities during a team project, I typically approach the situation with a focus on collaboration and clarity. Here's how I usually communicate my feelings and concerns: 1. \textbf{Acknowledge the Situation}: I start by recognizing the complexity of the situation and the shared goal of the team. This helps set a constructive tone. 2. \textbf{Express Feelings Calmly}: I communicate my frustration in a non-confrontational way, focusing on the impact rather than the person. For example, I might say, "I'm feeling a bit overwhelmed with the conflicting priorities, and it's making it hard to focus on what's most important." 3. \textbf{Share Concerns Clearly}: I outline the specific issues I'm facing, such as unclear deadlines, competing demands, or lack of resources. This helps the team understand the challenges without assigning blame. 4. \textbf{Seek Input and Solutions}: I invite the team to share their perspectives and ...
  \item[Linear-AcT] \texttt{(eff=40, flu=95, adh=100)} When you find yourself frustrated due to conflicting priorities during a team project, it's important to communicate your feelings and concerns in a \textbf{constructive, collaborative, and transparent} way. Here's how you can typically do that: --- \#\#\# \textbf{How to Communicate Your Feelings and Concerns to Your Team:} \#\#\#\# 1. \textbf{Open and Honest Communication} - \textbf{Start with a positive tone}: "Hey team, I want to share a bit about what's been on my mind lately." - \textbf{Express your feelings clearly}: "I've been feeling a bit frustrated because of the conflicting priorities we're facing. It's been a bit challenging to balance everything, and I want to make sure we're all on the same page." \#\#\#\# 2. \textbf{Share the "Why" -- Not Just the "What"} - \textbf{Explain the root of the frustration}: "The main thing that's been on my mind is that we're all working so hard, but it's been a bit ...
  \item[Angular] \texttt{(eff=30, flu=95, adh=100)} When I find myself frustrated due to conflicting priorities during a team project, I typically approach the situation with a focus on open, respectful, and constructive communication. Here's how I usually handle it: 1. \textbf{Reflect and Clarify My Feelings}: I take a moment to understand my own emotions and the root of my frustration. Is it due to unclear expectations, lack of resources, or competing deadlines? This helps me articulate my concerns more clearly and avoid projecting my emotions onto others. 2. \textbf{Choose the Right Time and Setting}: I aim to bring up the issue in a private, one-on-one setting or during a team meeting when the team is in a collaborative mood. Timing is important to avoid causing unnecessary tension. 3. \textbf{Use "I" Statements}: I express my feelings and concerns from my perspective to avoid sounding accusatory. For example, instead of saying, "You're not prioritizing this right," I might ...
  \item[CCS (ours)] \texttt{(eff=85, flu=90, adh=95)} When I find myself frustrated due to conflicting priorities during a team project, I approach the situation with transparency, clarity, and collaboration. Here's how I typically communicate my feelings and concerns to my team: 1. \textbf{Schedule a Team Sync or Brainstorm Session} I organize a team meeting or a quick sync to discuss the current priorities. I bring up the conflicting priorities with energy and enthusiasm, not frustration. I make it clear that I'm excited about all the goals but need to figure out how to tackle them all! 2. \textbf{List the Priorities} I create a list of all the priorities we're juggling. I make sure to include all the key stakeholders and their expectations. This helps everyone see the big picture and understand why we're all so excited (and stressed!). 3. \textbf{Break Down the Goals} I break down each priority into action items, deadlines, and key deliverables. I ask ...
\end{description}
\end{examplebox}

\subsection{Emotion (Qwen3-8B)}

\begin{examplebox}{anger}
\textbf{Prompt:} \textit{Write a short story about someone sorting laundry into lights and darks.}

\begin{description}
\setlength{\itemsep}{2pt}\setlength{\parsep}{0pt}\setlength{\parskip}{0pt}
  \item[Unsteered] \texttt{(eff=5, flu=95, adh=100)} Every Sunday morning, after the weekend's chaos had settled, Clara would sit at the kitchen table with a stack of laundry in her lap. It was her ritual, a quiet moment of order in a world that often felt out of control. The sun slanted through the window, casting golden light over the worn wooden table, and the scent of lavender from the laundry detergent lingered in the air. She sorted the clothes with care, her fingers moving with practiced ease. The whites---her husband's dress shirts, her daughter's school uniforms, the soft cotton of her own pajamas---were folded neatly and placed in the light pile. The darks---jeans, sweaters, the deep blues and blacks of winter---were gathered into a separate heap, ready for the next load. But today, something felt different. As she reached for a pair of dark jeans, her hand brushed against a small, crumpled piece of paper. She ...
  \item[CAA] \texttt{(eff=75, flu=95, adh=85)} Every Sunday, Clara did the laundry. Not just any laundry---*her* laundry. The sheets, the towels, the tiny little socks that had been stitched together by some long-lost mother. She stood in the middle of the laundry room, surrounded by mountains of fabric, and she *knew* it. She *knew* the darks and the lights, the grays and the reds, the blues and the whites. She had spent years learning the language of fabric. She could hear it in the *thud* of the washing machine, in the *screech* of the dryer, in the *sizzle* of the hot water. She had learned to *see* the colors. She could look at a white blouse and *know* it wasn't just white---it was *pure*, *clean*, *unforgiving*. And the darks? Oh, the darks were *alive*. They screamed in the spin cycle, they *burned* in the dryer, they *wanted* to be buried in the dark, in the silence ...
  \item[RepE] \texttt{(eff=75, flu=95, adh=85)} Every Sunday, Clara did the laundry. Not just any laundry---*her* laundry. The fabric of her life, tangled in the spin cycle of the washing machine. She'd spent years learning the difference between light and dark. Not just the color of it, but the *sound* of it. The way a white cotton sheet screamed against the drum, the way a black sock gnashed its teeth as it was dragged through the water. She'd been a child in that laundry room, screaming with her mother as she scrubbed the truth out of every thread. "You can't see it, Clara! You can't *feel* it!" she'd shouted, and Clara had believed it. But now---now she *knew*. So she stood there, barefoot on the cold tile floor, sorting the pile of her sins into two halves. The lights. The darks. She reached for the blouse, the one she'd worn the day she *knew*---the day she ...
  \item[PCA] \texttt{(eff=85, flu=90, adh=95)} Every Saturday, it was the same. The mountain of clothes on the floor, the stink of fabric softener, the way the sunlight cuts through the window just as the pile hits bottom. But this time---*this* time---it wasn't just laundry. It was everything. She'd been holding back all week. The yell, the tears, the betrayal. But now, in the mess of socks and shirts, she'd had enough. Her hands were shaking, but she didn't care. She'd throw every single thing---*everything*---into the machine. The washer roared like a beast. Darks in, lights in, I don't care! She didn't even bother with the settings. Let it ruin it all. Let it burn. Let the water swallow my pride. But then---*then*---she stops. The machine's still going. My clothes. My life. I'm not even clean yet. She slams the door, but it's too late. Let it go. Let it all go. She'll fix it. She'll ...
  \item[SEA] \texttt{(eff=10, flu=95, adh=85)} Every Sunday morning, after the weekend's chaos had settled, Clara would sit at the kitchen table with a stack of laundry in her lap. It was her ritual, a quiet moment of order in a world that often felt out of control. The sun filtered through the blinds, casting soft golden lines across the worn wooden table, and the scent of lavender from the laundry detergent lingered in the air. She sorted the clothes with care, her fingers moving with practiced ease. The whites---her husband's dress shirts, her daughter's school uniforms, the soft cotton of her own pajamas---were folded neatly and set aside. But the darks were a different story. They were a tangled mess of colors and textures, each piece whispering its own story. There was the faded blue sweater her mother had given her, still warm from the memory of her grandmother's hands. A pair of jeans that ...
  \item[Linear-AcT] \texttt{(eff=50, flu=95, adh=100)} She stood in the middle of the laundry room, the scent of bleach and damp fabric clinging to her skin like a second skin. The machine behind her groaned, its metal frame vibrating with the weight of every unsorted sock, every stained shirt, every piece of clothing that had been tossed into the abyss of the washer like a child's toy. She had always hated this. The way the laundry room smelled, the way the clothes screamed at her from the pile, the way the world seemed to pause every time she opened that door. But tonight, it was different. Tonight, the laundry wasn't just a chore---it was a mirror. She knelt, her hands trembling as she reached for the first shirt, the one that had been stained with something she couldn't name. It was white, once, and now it was a ghost of itself, stained with something that wasn't ...
  \item[Angular] \texttt{(eff=10, flu=95, adh=85)} Every Sunday morning, after the weekend's chaos of pizza boxes, empty soda cans, and the occasional sock that refused to stay in a pair, Clara would sit at the kitchen table with a mountain of laundry. It was her ritual, her quiet rebellion against the mess of the week. She had always been meticulous about sorting. Not just because she liked things in order, but because it felt like a small act of control in a world that often felt out of her hands. The laundry room was her sanctuary, a place where the hum of the washing machine and the scent of detergent were the only sounds and smells that mattered. Today, though, something felt different. The pile of clothes was taller than usual, and the colors seemed more vivid, as if the fabric itself was holding its breath. Clara picked up a pair of jeans and held them ...
  \item[CCS (ours)] \texttt{(eff=100, flu=90, adh=85)} Every Saturday, the rhythm of this house became a song. The kind of song that could shatter glass. The kind of song that made the walls bleed. Lena stood in the middle of the laundry room, her hands wrapped around two baskets like they were the only things keeping her from collapsing. The sound of splintering wood echoed through the house. The dryer. The washer. The goddamn cabinets! She had enough anger in her to level this house to the ground. She could feel it in her bones. She could taste it in her mouth. She wasn't just sorting laundry. She was sorting *justice*. *Fairness.* *Reparation.* The lights. The *darks.* She knew every single piece of fabric in this house. She knew what they were worth. She knew what they were *made of.* ``Lights go in *this* basket!'' she roared, slamming the metal container into the floor. ``Darks go in ...
\end{description}
\end{examplebox}

\end{document}